\documentclass[twoside,11pt]{article}
\usepackage{epsfig, amsmath, subfigure, soul, color}
\pdfoutput=1 
\usepackage[longnamesfirst]{natbib}
\usepackage{jmlr2e}
\usepackage{pslatex}
\usepackage[small]{caption}
\usepackage{placeins}
\usepackage{floatpag}
\usepackage{tabularx}

\definecolor{WowColor}{rgb}{.75,0,.75}
\definecolor{SubtleColor}{rgb}{0,0,.50}
\definecolor{EvenSubtlerColor}{rgb}{0.2,0,.40}

\newcommand{\TBD}[1]{\textcolor{SubtleColor}{ {\tiny \bf (!)} #1}}

\newcounter{margincounter}
\newcommand{\displaycounter}{{\arabic{margincounter}}}
\newcommand{\incdisplaycounter}{{\stepcounter{margincounter}\arabic{margincounter}}}

\newcommand{\fTBD}[1]{\textcolor{SubtleColor}{$\,^{(\incdisplaycounter)}$}\marginpar{\tiny\textcolor{SubtleColor}{ {\tiny $(\displaycounter)$} #1}}}

\definecolor{LightGray}{rgb}{.6,.6,.6}

\newcommand*{\bigCI}{%
  \mathrel{\text{%
    {\rotatebox[origin=c]{90}{\resizebox{2.25ex}{1.65ex}{$\vDash$}}}%
  }}%
}
\definecolor{darkred}{rgb}{0.5,0,0}
\definecolor{darkgreen}{rgb}{0, 0.3,0}
\definecolor{darkblue}{rgb}{0,0,0.6}

\newcommand{\given}{\,\!\,|\,\!\,}

\def\[#1\]{\begin{align}#1\end{align}}

\newcommand{\HIDE}[1]{}

\definecolor{MyGreen}{rgb}{.75,0,.75}
\definecolor{RealGreen}{rgb}{0,1,0}
\definecolor{DarkGreen}{rgb}{0.1,0.4,0.1}
\definecolor{ActualGreen}{rgb}{0.0,0.5,0.0}
\definecolor{MyBlue}{rgb}{0,0,1}
\definecolor{MyPurple}{rgb}{0.6,0,0.6}
\definecolor{MyRed}{rgb}{1,0,0}

\newcommand{\cef}[1]{\textcolor{DarkGreen}{ ({\bf CEF:\ } #1)}}
\newcommand{\vkm}[1]{\textcolor{MyRed}{ ({\bf VKM:\ } #1)}}
\newcommand{\dmr}[1]{\textcolor{MyPurple}{ ({\bf DMR:\ } #1)}}
\newcommand{\both}[1]{\textcolor{ActualGreen}{ ({\bf Both:\ } #1)}}
\newcommand{\low}[1]{\textcolor{MyBlue}{ ({\bf Low Priority:\ } #1)}}

\newcommand{\fvkmlook}[1]{\textcolor{MyRed}{$\,^{(\incdisplaycounter)}$}\marginpar{\tiny\textcolor{MyRed}{ {\tiny CEF$\rightarrow$VKM: $(\displaycounter)$} #1}}}

\newcommand{\fcef}[1]{\textcolor{DarkGreen}{$\,^{(\incdisplaycounter)}$}\marginpar{\tiny\textcolor{DarkGreen}{ {\tiny CEF: $(\displaycounter)$} #1}}}

\renewcommand{\cef}[1]{}
\renewcommand{\vkm}[1]{}
\renewcommand{\dmr}[1]{}
\renewcommand{\both}[1]{}
\renewcommand{\low}[1]{}
\renewcommand{\TBD}[1]{}
\renewcommand{\fTBD}[1]{}
\renewcommand{\fvkmlook}[1]{}
\renewcommand{\fcef}[1]{}

\begin{document}

\title{CrossCat: A Fully Bayesian Nonparametric Method for Analyzing Heterogeneous, High Dimensional Data}

\author{\name Vikash Mansinghka \email vkm@mit.edu \\
\name Patrick Shafto \email p.shafto@louisville.edu \\
\name Eric Jonas \email jonas@priorknowledge.net \\
\name Cap Petschulat \email cap@priorknowledge.net \\
\name Max Gasner \email max@priorknowledge.net \\
\name Joshua B. Tenenbaum \email jbt@mit.edu}

\editor{David Blei}

\maketitle

\begin{abstract}There is a widespread need for statistical methods
  that can analyze high-dimensional datasets without imposing
  restrictive or opaque modeling assumptions. This paper describes a
  domain-general data analysis method called CrossCat. CrossCat infers
  multiple non-overlapping views of the data, each consisting of a
  subset of the variables, and uses a separate nonparametric mixture
  to model each view. CrossCat is based on approximately Bayesian
  inference in a hierarchical, nonparametric model for data
  tables. This model consists of a Dirichlet process mixture over the
  columns of a data table in which each mixture component is itself an
  independent Dirichlet process mixture over the rows; the inner
  mixture components are simple parametric models whose form depends
  on the types of data in the table. CrossCat combines strengths of
  mixture modeling and Bayesian network structure learning. Like
  mixture modeling, CrossCat can model a broad class of distributions
  by positing latent variables, and produces representations that can
  be efficiently conditioned and sampled from for prediction. Like
  Bayesian networks, CrossCat represents the dependencies and
  independencies between variables, and thus remains accurate when
  there are multiple statistical signals. Inference is done via a
  scalable Gibbs sampling scheme; this paper shows that it works well
  in practice. This paper also includes empirical results on
  heterogeneous tabular data of up to 10 million cells, such as
  hospital cost and quality measures, voting records, unemployment
  rates, gene expression measurements, and images of handwritten
  digits. CrossCat infers structure that is consistent with accepted
  findings and common-sense knowledge in multiple domains and yields
  predictive accuracy competitive with generative, discriminative, and
  model-free alternatives.
\end{abstract}

\begin{keywords}
  Bayesian nonparametrics, Dirichlet processes, Markov chain Monte Carlo, multivariate analysis, structure learning, unsupervised learning, semi-supervised learning
\end{keywords}

\thanks{{\bf Acknowledgments:} The authors thank Kevin Murphy, Ryan
  Rifkin, Cameron Freer, Daniel Roy, Bill Lazarus, David Jensen, Beau
  Cronin, Rax Dillon, and the anonymous reviewers and editor for
  valuable suggestions. This work was supported in part by gifts from
  Google, NTT Communication Sciences Laboratory, and Eli Lilly \& Co;
  by the Army Research Office under agreement number W911NF-13-1-0212;
  and by DARPA via the XDATA and PPAML programs. Any opinions,
  findings, and conclusions or recommendations expressed in this work
  are those of the authors and do not necessarily reflect the views of
  any of the above sponsors.}

\section{Introduction}

High-dimensional datasets containing data of multiple types have
become commonplace. These datasets are often represented as tables,
where rows correspond to data vectors, columns correspond to
observable variables or features, and the whole table is treated as a
random subsample from a statistical population
\citep{hastie2005elements}. This setting brings new opportunities as
well as new statistical challenges \citep{napmassive,
  wassermanlow}. In principle, the dimensionality and coverage of some
of these datasets is sufficient to rigorously answer to fine-grained
questions about small sub-populations. This size and richness also
enables the detection of subtle predictive relationships, including
those that depend on aggregating individually weak signals from large
numbers of variables. Challenges include integrating data of
heterogeneous types \citep{napmassive}, suppressing spurious patterns
\citep{benjamini1995controlling, attia2009use}, selecting features
\citep{wassermanlow, weston2001feature}, and the prevalence of
non-ignorable missing data.

This paper describes CrossCat, a general-purpose Bayesian method for
analyzing high-dimensional mixed-type datasets that aims to mitigate
these challenges. CrossCat is based on approximate inference in a
hierarchical, nonparametric Bayesian model. This model is comprised of
an ``outer'' Dirichlet process mixture over the columns of a table,
with components that are themselves independent ``inner'' Dirichlet
process mixture models over the rows. CrossCat is parameterized on a
per-table basis by data type specific component models --- for
example, Beta-Bernoulli models for binary values and Normal-Gamma
models for numerical values. Each ``inner'' mixture is solely
responsible for modeling a subset of the variables. Each hypothesis
assumes a specific set of marginal dependencies and
independencies. This formulation supports scalable algorithms for
learning and prediction, specifically a collapsed MCMC scheme that
marginalizes out all but the latent discrete state and hyper
parameters.
 
\begin{figure}
\begin{center}
\vspace{-1in}
\includegraphics[width=6.5in]{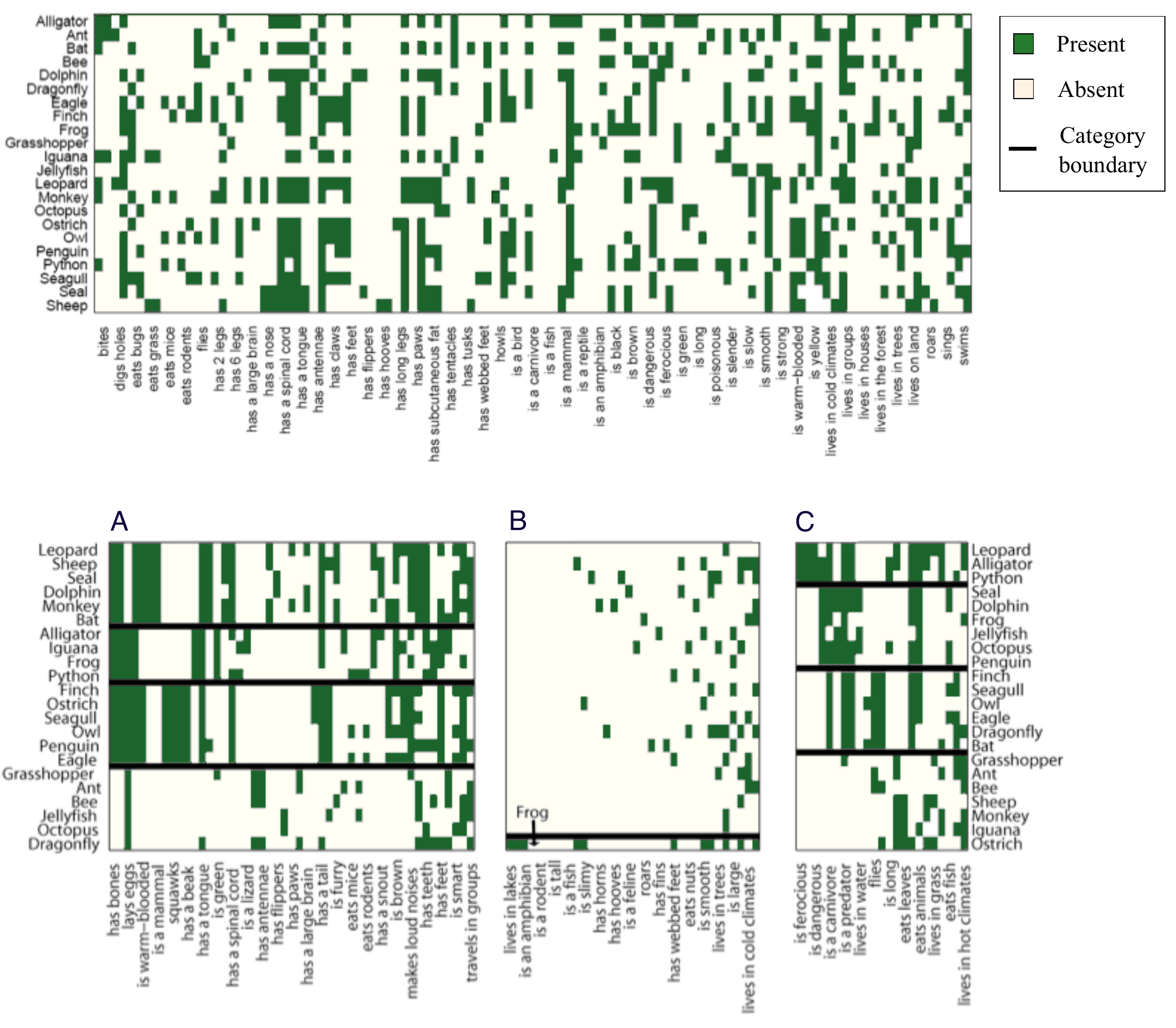}
\end{center}
\vspace{-0.2in}
\caption{ {\bf An illustration of the latent structure posited by
    cross-categorization on a dataset of common-sense judgments about
    animals.}  The figure shows the raw input data and one posterior
  sample from a dataset containing animals and their
  properties. CrossCat finds three independent signals, or {\em
    views}. A taxonomic clustering (left, A) comprises groups of mammals,
  amphibians and reptiles, birds, and invertebrates, and explains
  primarily anatomical and physiological variables.  An ecological
  clustering (right, C) cross-cuts the taxonomic groups and specifies
  groups of carnivorous land predators, sea animals, flying animals,
  and other land animals, and explains primarily habitat and behavior
  variables.  Finally all animals (except frogs) are lumped together
  into a cluster of miscellaneous features (center, B) that accounts for
  a set of idiosyncratic or sparse ``noise'' variables.}
\label{figure:crosscat-schematic}
\end{figure}

The name ``CrossCat'' is derived from the combinatorial skeleton of
this probabilistic model. Each approximate posterior sample represents
a {\em cross-categorization} of the input data table. In a
cross-categorization, the variables are partitioned into a set of {\em
  views}, with a separate partition of the entities into {\em
  categories} with respect to the variables in each {\em view}. Each
$(category, variable)$ pair contains the sufficient statistics or
latent state needed by its associated component model. See Figure
\ref{figure:crosscat-schematic} for an illustration of this
structure. From the standpoint of structure learning, CrossCat finds
multiple, cross-cutting categorizations or clusterings of the data
table. Each non-overlapping system of categories is context-sensitive
in that it explains a different subset of the variables. Conditional
densities are straightforward to calculate and to sample from. Doing
so first requires dividing the conditioning and target variables into
views, then sampling a category for each view. The distribution on
categories must reflect the values of the conditioning
variables. After choosing a category it is straightforward to sample
predictions or evaluate predictive densities for each target variable
by using the appropriate component model.

Standard approaches for inferring representations for joint
distributions from data, such as Bayesian networks, mixture models,
and sparse multivariate Gaussians, each exhibit complementary
strengths and limitations. Each method exhibits distinct strengths and
weaknesses:

\begin{enumerate}

\item {\bf Bayesian networks and structure learning.} \\

  The main advantage offered by Bayesian networks in this setting is
  that they can use a separate network to model each group of
  variables. From a statistical perspective, Bayes nets may be
  effective for sufficiently large, purely discrete datasets where all
  variables are observed and no hidden variables are needed to
  accurately model the true data generator. The core modeling
  difficulty is that many relevant joint distributions are either
  impossible to represent using a Bayesian network or require
  prohibitively complex parameterizations. For example, without hidden
  variables, Bayes nets must emulate the effect of any hidden
  variables by implicitly marginalizing them out, yielding a dense set
  of connections. These artificial edges can reduce statistical
  efficiency: each new parent for a given node can multiplicatively
  increase the number of parameters to estimate
  \citep{elidan2001discovering, elidan2005learning}. There are also
  computational difficulties. First, there are no known scalable
  techniques for fully Bayesian learning of Bayesian networks, so
  posterior uncertainty about the dependence structure is lost.
  Second, even when the training data is fully observed, i.e. all
  variables are observed, search through the space of networks is
  computationally demanding. Third, if the data is incomplete (or
  hidden variables are posited), a complex inference subproblem needs
  to be solved in the inner loop of structure search.

\item {\bf Parametric and semi-parametric mixture modeling.} \\

  Mixtures of simple parametric models have several appealing
  properties in this setting. First, they can accurately emulate the
  joint distribution within each group of variables by introducing a
  sufficiently large number of mixture components. Second,
  heterogeneous data types are naturally handled using independent
  parametric models for each variable chosen based on the type of data
  it contains. Third, learning and prediction can both be done via
  MCMC techniques that are linear time per iteration, with constant
  factors and iteration counts that are often acceptable in
  practice. Unfortunately, mixture models assume that all variables
  are (marginally) coupled through the latent mixture component
  assignments. As a result, posterior samples will often contain good
  categorizations for one group of variables, but these same
  categories treat all other groups of mutually dependent variables as
  noise. This can lead to dramatic under-fitting in high dimensions or
  when missing values are frequent; this paper includes experiments
  illustrating this failure mode. Thus if the total number of
  variables is small enough, and the natural cluster structure of all
  groups of variables is sufficiently similar, and there is enough
  data, mixture models may perform well.

\item {\bf Multivariate Gaussians with sparse inverse covariances.} \\

  High-dimensional continuous distributions are often modeled as
  multivariate Gaussians with sparse conditional dependencies
  \citep{meinshausen2006high}. Several parameter estimation techniques
  are available; see e.g. \citep{friedman2008sparse}. The pairwise
  dependencies produced by these methods form an undirected graph. The
  underlying assumptions are most appropriate when the number of
  variables and observations are sufficiently large, the data is
  naturally continuous and fully observed, and the joint distribution
  is approximately unimodal. A key advantage of these methods is the
  availability of fast algorithms for parameter estimation (though
  extensions for handling missing values require solving challenging
  non-convex optimization problems \citep{stadler2012missing}). These
  methods also have two main limitations. First, the assumption of
  joint Gaussianity is unrealistic in many situations
  \citep{wassermanlow}. Second, discrete values must be transformed
  into numerical values; this invalidates estimates of predictive
  uncertainty, and can generate other surprising behaviors.

\end{enumerate}

CrossCat combines key computational and statistical strengths of each
of these methods. As with nonparametric mixture modeling, CrossCat
admits a scalable MCMC algorithm for posterior sampling, handles
incomplete data, and does not impose restrictions on data
types. CrossCat also preserves the asymptotic consistency of density
estimation via Dirichlet process mixture modeling \citep{Dunson09},
and can emulate a broad class of generative processes and joint
distributions given enough data. However, unlike mixture modeling but
like Bayesian network structure learning, CrossCat can also detect
independencies between variables. The ``outer'' Dirichlet process
mixture partitions variables into groups that are independent of one
another. As with estimation of sparse multivariate Gaussians (but
unlike Bayesian network modeling), CrossCat can handle complex
continuous distributions and report pairwise measures of association
between variables. However, in CrossCat, the couplings between
variables can be nonlinear and heteroscedastic, and induce complex,
multi-modal distributions. These statistical properties are
illustrated using synthetic tests designed to strain the CrossCat
modeling assumptions and inference algorithm.

This paper illustrates the flexibility of CrossCat by applying it to
several exploratory analysis and predictive modeling tasks. Results on
several real-world datasets of up to 10 million cells are
described. Examples include measures of hospital cost and quality,
voting records, US state-level unemployment time series, and
handwritten digit images. These experiments show that CrossCat can
extract latent structures that are consistent with accepted findings
and common-sense knowledge in multiple domains. They also show that
CrossCat can yield favorable predictive accuracy as compared to
generative, discriminative, and model-free baselines.

The remainder of this paper is organized as follows. This section
concludes with a discussion of related work. Section 2 focuses on
generative model and approximate inference scheme behind
CrossCat. Section 3 describes empirical results, and section 4
contains a broad discussion and summary of contributions.

\subsection{Related Work}

The observation that multiple alternative clusterings can often
explain data better than a single clustering is not new to this
paper. Methods for finding multiple clusterings have been developed in
several fields, including by the authors of this paper \citep[see
e.g.][]{niu2010multiple,cui2007non,guanEtal10,liShafto11,rodriguezEtal98,shaftoetal06,shaftoEtal11_crosscat,rossz06}. For
example, \citet{rossz06} used an EM approach to fit a parametric
mixture of mixtures and applied it to image modeling.  As
nonparametric mixtures and model selection over finite mixtures can
behave similarly, it might seem that a nonparametric formulation is a
small modification. In fact, nonparametric formulation presented here
is based on a super-exponentially larger space of model complexities
that includes all possible numbers and sizes of views, and for each
view, all possible numbers and sizes of categories. This
expressiveness is necessary for the broad applicability of
CrossCat. Cross-validation over this set is intractable, motivating
the nonparametric formulation and sampling scheme used in this paper.

It is instructive to compare and contrast CrossCat with related
hierarchical Bayesian models that link multiple Dirichlet process
mixture models, such as the nested Dirichlet process
\citep{rodriguez2008nested} and the hierarchical Dirichlet process
\citep{teh2006hierarchical}. See \citet{jordan2010hierarchical} for a
thorough review. This section contains a brief discussion of the most
important similarities and differences. The hierarchical Dirichlet
process applies independent Dirichlet processes to each dataset, whose
atoms are themselves draws from a single shared Dirichlet process. It
thus enables a single pool of clusters to be shared and sparsely
expressed in otherwise independent clusterings of several
datasets. The differences are substantial. For example, with the
hierarchical Dirichlet process, the number of Dirichlet processes is
fixed in advance. In CrossCat, each atom on one Dirichlet process is
associated with its own Dirichlet process, and inference is used to
determine the number that will be expressed in a given finite
dataset. 

The nested Dirichlet process shares this combinatorial structure with
CrossCat, but has been used to build very different statistical
models. \citep{rodriguez2008nested} introduces it as a model for
multiple related datasets. The model consists of a Dirichlet process
mixture over datasets where each component is another Dirichlet
process mixture models over the items in that dataset. From a
statistical perspective, it can be helpful to think of this
construction as follows. First, a top-level Dirichlet process is used
to cluster datasets. Second, all datasets in the same cluster are
pooled and their contents are modeled via a single clustering,
provided by the lower-level Dirichlet process mixture model associated
with that dataset cluster.

The differences between CrossCat and the nested Dirichlet process are
clearest in terms of the nested Chinese restaurant process
representation of the nested DP \citep{griffiths2004hierarchical,
  blei2010nested}. In a 2-layer nested Chinese restaurant process,
there is one customer per data vector. Each customer starts at the top
level, sits a table at their current level according to a CRP, and
descends to the CRP at the level below that the chosen table
contains. In CrossCat, the top level CRP partitions the variables into
views, and the lower level CRPs partition the data vectors into
categories for each view. If there are $K$ tables in top CRP, i.e. the
dataset is divided into $K$ views, then adding one datapoint leads to
the seating of $K$ new customers at level 2. Each of these customers
is deterministically assigned to a distinct table. Also, whenever a
new customer is created at the top restaurant, in addition to creating
a new CRP at the level below, $R$ customers are immediately seated
below it (one per row in the dataset).

Close relatives of CrossCat have been introduced by the authors of
this paper in the cognitive science literature, and also by other
authors in machine learning and statistics. This paper goes beyond
this previous work in several ways. \citet{guanEtal10} uses a
variational algorithm for inference, while \citet{rodriguezEtal98}
uses a sampler for the stick breaking representation for a Pitman-Yor
(as opposed to Dirichlet Process) variant of the model. CrossCat is
instead based on samplers that (i) operate over the combinatorial
(i.e. Chinese restaurant) representation of the model, not the
stick-breaking representation, and (ii) perform fully Bayesian
inference over all hyper-parameters. This formulation leads to
CrossCat's scalability and robustness. This paper includes results on
tables with millions of cells, without any parameter tuning, in
contrast to the 148x500 gene expression subsample analyzed in
\citet{rodriguezEtal98}. These other papers include empirical results
comparable in size to the authors' experiments from
\citet{shaftoetal06} and \citet{mansinghkaEtal09}; these are 10-100x
smaller than some of the examples from this paper. Additionally, all
the previous work on variants of the CrossCat model focused on
clustering, and did not articulate its use as a general model for
high-dimensional data generators. For example, \citet{guanEtal10} does
not include predictions, although \citet{rodriguezEtal98} does discuss
an example of imputation on a 51x26 table.

To the best of our knowledge, this paper is the first to introduce a
fully Bayesian, domain-general, semi-parametric method for estimating
the joint distributions of high-dimensional data. This method appears
to be the only joint density estimation technique that simultaneously
supports heterogeneous data types, detects independencies, and
produces representations that support efficient prediction. This paper
is also the first to empirically demonstrate the effectiveness of the
underlying probabilistic model and inference algorithm on multiple
real-world datasets with mixed types, and the first to compare
predictions and latent structures from this kind of model against
multiple generative, discriminative and model-free baselines.

\section{The CrossCat Model and Inference Algorithm}

CrossCat is based on inference in a column-wise Dirichlet process
mixture of Dirichlet process mixture models
\citep{escobarWest95,rasmussen00} over the rows. The ``outer'' or
``column-wise'' Dirichlet process mixture determines which
dimensions/variables should be modeled together at all, and which
should be modeled independently. The ``inner'' or ``row-wise''
mixtures are used to summarize the joint distribution of each group of
dimensions/variables that are stochastically assigned to the same
modeling subproblem.

This paper presents the Dirichlet processes in CrossCat via the
convenient Chinese restaurant process representation
\citep{pitman96}. Recall that the Dirichlet process is a stochastic
process that maps an arbitrary underlying base measure into a measure
over discrete atoms, where each atom is associated with a single draw
from the base measure. In a set of repeated draws from this discrete
measure, some atoms are likely to occur multiple times. In
nonparametric Bayesian mixture modeling, each atom corresponds to a
set of parameters for some mixture component; ``popular'' atoms
correspond to mixture components with high weight. The Chinese
restaurant process is a stochastic process that corresponds to the
discrete residue of the Dirichlet process. It is sequential, easy to
describe, easy to simulate, and exchangeable. It is often used to
represent nonparametric mixture models as follows. Each data item is
viewed as a customer at a restaurant with an infinite number of
tables. Each table corresponds to a mixture component; the customers
at each table thus comprise the groups of data that are modeled by the
same mixture component.  The choice probabilities follow a simple
``rich-gets-richer'' scheme. Let $m_j$ be the number of customers
(data items) seated at a given table $j$, and $z_i$ be the table
assignment of customer $i$ (with the first table $z_0$ = 0), then the
conditional probability distribution governing the Chinese restaurant
process with concentration parameter $\alpha$ is:

$$
Pr(z_i = j) \propto \left\{
\begin{array}{ll}
\alpha & \mathrm{if} \hspace{1mm} j = \mathrm{max}(\vec{z})+1 \\
m_j & o.w. \\
\end{array}\right.
$$

This sequence of conditional probabilities induces a distribution over
the partitions of the data that is equivalent to the marginal
distribution on equivalence classes of atom assignments under the
Dirichlet process. The Chinese restaurant process provides a simple
but flexible modeling tool: the number of components in a mixture can
be determined by the data, with support over all logically possible
clusterings. In CrossCat, the {\em number} of Chinese restaurant
processes (over the rows) is determined by the number of tables in a
Chinese restaurant process over the columns. The data itself is
modeled by datatype-specific component models for each dimension
(column) of the target table.

\subsection{The Generative Process}

The generative process behind CrossCat unfolds in three
steps:

\begin{enumerate}
\item {\bf Generating hyper-parameters and latent structure.} First, the
  hyper-parameters $\vec{\lambda}_d$ for the component models for each
  dimension are chosen from a vague hyper-prior $V_d$ that is
  appropriate\footnote{The hyper-prior must only assign positive
    density to valid hyper-parameter values and be sufficiently broad
    for the marginal distribution for a single data cell has
    comparable spread to the actual data being analyzed. We have
    explored multiple hyper-priors that satisfy these constraints on
    analyses similar to those from this paper; there was little
    apparent variation. Examples for strictly positive, real-valued
    hyper-parameters include vague $\mathrm{Gamma}(k=1, \theta=1)$
    hyper-prior, uniform priors over a broad range, and both linear
    and logarithmic discretizations. Our reference implementation
    uses a set of simple data-dependent heuristics to determine
    sufficiently broad ranges. Chinese restaurant process
    concentration parameters are given 100-bin log-scale grid discrete
    priors; concentration parameters for the finite-dimensional
    Dirichlet distributions used to generate component parameters for discrete data
    have the same form. For Normal-Gamma models,
    $\mathrm{min}(\vec{x}_{(\cdot, d)}) \le \mu_d \le
    \mathrm{max}(\vec{x}_{(\cdot, d)})$.} for the type of data in
  $d$. Second, the concentration parameter $\alpha$ for the outer
  Chinese restaurant process is sampled from a vague gamma
  hyper-prior. Third, a partition of the variables into views,
  $\vec{z}$, is sampled from this outer Chinese restaurant
  process. Fourth, for each view, $v \in \vec{z}$, a concentration
  parameter $\alpha_v$ is sampled from a vague hyper-prior. Fifth, for
  each view $v$, a partition of the rows $\vec{y}^v$ is drawn using
  the appropriate inner Chinese restaurant process with concentration
  $\alpha_v$.

\item {\bf Generating category parameters for uncollapsed variables.}
  This paper uses $u_d$ as an indicator of whether a given
  variable/dimension $d$ is uncollapsed ($u_d = 1$) or collapsed ($u_d
  = 0$). For each uncollapsed variable, parameters $\vec{\theta_c^d}$
  must be generated for each category $c$ from a datatype-compatible
  prior model $M_d$.

\item {\bf Generating the observed data given hyper-parameters, latent
    structure, and parameters.} The dataset $\mathbf{X} =
  \{x_{(r,d)}\}$ is generated separately for each variable $d$ and for
  each category $c \in \vec{y}^v$ in the view $v = z_d$ for that
  variable. For uncollapsed dimensions, this is done by repeatedly
  simulating from a likelihood model $L_d$. For collapsed dimensions,
  we use an exchangeably coupled model $ML_d$ to generate all the data
  in each category at once.
\end{enumerate}

\noindent The details of the CrossCat generative process are as follows:

\begin{enumerate}

	\item Generate $\alpha_D$, the concentration hyper-parameter for the Chinese Restaurant Process over dimensions, from a generic Gamma hyper-prior: $\alpha_D \sim \mathrm{Gamma}(k = 1, \theta = 1)$. 
		
	\item For each dimension $d \in D$:
	
	\begin{enumerate}
	
		\item Generate hyper-parameters $\vec{\lambda}_d$ from a data type appropriate hyper-prior with density $p(\vec{\lambda}_d) = V_d(\vec{\lambda}_d)$, as described above. Binary data is handled by an asymmetric Beta-Bernoulli model with pseudocounts $\vec{\lambda}_d = [\alpha_d, \beta_d]$. Discrete data is handled by a symmetric Dirichlet-Discrete model with concentration parameter $\lambda_d$. Continuous data is handled by a Normal-Gamma model with $\vec{\lambda}_d = (\mu_d , \kappa_d, \upsilon_d, \tau_d)$, where $\mu_d$ is the mean, $\kappa_d$ is the effective number of observations, $\upsilon_d$ is the degrees of freedom, and $\tau_d$ is the sum of squares. 

		\item Assign dimension $d$ to a view $z_d$ from a Chinese Restaurant Process with concentration hyper-parameter $\alpha_D$, conditional on all previous draws: $z_d \sim \mathrm{CRP}(\{z_0, \cdots, z_{d-1}\}; \alpha_D)$
	
	\end{enumerate}
	
	\item For each view $v$ in the dimension partition $\vec{z}$:
		\begin{enumerate}

		\item Generate $\alpha_v$, the concentration hyper-parameter for the Chinese Restaurant Process over categories in view $v$, from a generic hyper-prior: $\alpha_v \sim \mathrm{Gamma}(k = 1, \theta = 1)$.
		
		\item For each observed data point (i.e. row of the table) $r \in R$, generate a category assignment $y^v_r$ from a Chinese Restaurant Process with concentration parameter $\alpha_v$, conditional on all previous draws: $y^v_r \sim \mathrm{CRP}(\{y^v_0, \cdots, y^v_{r-1}\}; \alpha_v)$

		\item For each category $c$ in the row partition for this view $\vec{y}^v$:

		\begin{enumerate}

			\item For each dimension $d$ such that $u_d = 1$ (i.e. its component models are uncollapsed), generate component model parameters
			  $\vec{\theta_c^d}$ from the appropriate prior with density $M_d(\cdot; \vec{\lambda}_d)$ using hyper-parameters $\vec{\lambda}_d$, as follows:
			  
			  \begin{enumerate}
			  
			  \item For binary data, we have a scalar $\theta_c^d$ equal to the probability that dimension $d$ is equal to 1 for rows from category $c$, drawn from a Beta distribution: $\theta_c^d \sim \mathrm{Beta}(\alpha_d, \beta_d)$, where values from the hyper-parameter vector $\vec{\lambda}_d = [\alpha_d, \beta_d]$.
			  
			  \item For categorical data, we have a vector-valued $\vec{\theta_c^d}$ of probabilities, drawn from a symmetric Dirichlet distribution with concentration parameter $\lambda_d$: $\vec{\theta_c^d} \sim \mathrm{Dirichlet}(\lambda_d)$.
			  
			  \item For continuous data, we have $\vec{\theta_c^d} = (\mu_c^d, \sigma_c^d)$, the mean and variance of the data in the component, drawn from a Normal-Gamma distribution $(\mu_c^d, \sigma_c^d) \sim \mathrm{NormalGamma}(\vec{\lambda}_d)$.
			  
			  \end{enumerate}
			  
			\item Let $\vec{x}_{(\cdot, d)}^c$ contain all $x_{(r,d)}$ in this component, i.e. for $r$ such that $y^{z_d}_r = c$. Generate the data in this component, as follows:

			\begin{enumerate}

                        \item If $u_d = 1$, i.e. $d$ is uncollapsed,
                          then generate each $x_{(r,d)}$ from the
                          appropriate likelihood model $L_d(\cdot;
                          \vec{\theta_c^d})$. For binary data, we have
                          $x_{(r, d)} \sim
                          \mathrm{Bernoulli}(\theta_c^d)$; for
                          categorical data, we have $x_{(r, d)}\sim
                          \mathrm{Multinomial}(\vec{\theta_c^d})$; for
                          continuous data, we have $x_{(r, d)} \sim
                          \mathrm{Normal}(\mu_c^d, \sigma_c^d)$.
			
				\item If $u_d = 0$, so $d$ is
                                  collapsed, generate the entire
                                  contents of $\vec{x}_{(\cdot, d)}^c$
                                  by directly simulating from the
                                  marginalized component model that
                                  with density $ML_d(\vec{x}_{(\cdot,
                                    d)}; \vec{\lambda}_d)$. One
                                  approach is to sample from the
                                  sequence of predictive distributions
                                  $P(x_{(r_i, d)} | \vec{x}_{(\cdot,
                                    d)}^{-r_i}. \vec{\lambda}_d)$,
                                  induced by $M_d$ and $L_d$, indexing
                                  over rows $r_i$ in c.
			 	
			\end{enumerate}

		\end{enumerate}

	\end{enumerate}

\end{enumerate}

The key steps in this process can be concisely described:

\begin{flalign*}
\alpha_D & \sim \mathrm{Gamma}(k = 1, \theta = 1) &&  \hfill \\
\vec{\lambda}_d & \sim \ V_d(\cdot) && \mathrm{for each}\ d \in \{1, \cdots, D\} \\
z_d & \sim \ \mathrm{CRP}(\{z_i \mid i \neq d\}; \alpha_D) && \mathrm{for each}\ d \in \{1, \cdots, D\}\\
\alpha_v & \sim \ \mathrm{Gamma}(k = 1, \theta = 1) && \mathrm{for each}\ v \in \vec{z} \\
y^v_r & \sim \ \mathrm{CRP}(\{y^v_i \mid i \neq r\}; \alpha_v) && \mathrm{for each}\ v \in \vec{z}\ \mathrm{and} \\ & && r \in \{1, \cdots, R\} \\
\vec{\theta_c^d} & \sim \ M_d(\cdot; \vec{\lambda}_d) && \mathrm{for each}\ v \in \vec{z},c \in \vec{y}^v,\ \mathrm{and}\ d\ \mathrm{such}\ \mathrm{that} \\ &  && z_d = v\ \mathrm{and}\ u_d = 1 \\
\vec{x}_{(\cdot, d)}^c = \{ x_{(r,d)} \mid y^{z_d}_r = c \} & \sim \begin{cases}
\prod_r L_d(\vec{\theta_c^d}) & \mathrm{if}\ u_d = 1 \\
ML_d(\vec{\lambda}_d) & \mathrm{if}\ u_d = 0 \\
\end{cases} 
 &&\mathrm{for each}\ v\in \vec{z}\ \mathrm{and}\ \mathrm{each}\ c \in \vec{y}^v
\end{flalign*}

\subsection{The joint probability density}

Recall that the following dataset-specific information is needed to fully the specify the CrossCat model:
\begin{enumerate}
\item $V_d(\cdot)$, a generic hyper-prior of the appropriate type for variable/dimension $d$.
\item $\{u_d\}$, the indicators for which variables are uncollapsed.
\item $M_d(\cdot)\ \mathrm{and}\ L_D(\cdot)\ \forall\ d\ s.t.\ u_d =
  1$, a datatype-appropriate parameter prior (e.g. a Beta prior for
  binary data, Normal-Gamma for continuous data, or Dirichlet for
  discrete data) and likelihood model (e.g.\ Bernoulli, Normal or
  Multinomial).
\item $ML_d(\cdot)\ \forall\ d\ s.t.\ u_d = 0$, a datatype-appropriate marginal likelihood model, e.g. the collapsed version of the conjugate pair formed
by some $M_d$ and $L_d$.
\item $T_d(\{x\})$, the sufficient statistics for the component model
  for some collapsed dimension $d$ from a subset of the data $\{x\}$.
  Arbitrary non-conjugate component models can be numerically
  collapsed by choosing $T_d(\{x\}) = \{x\}$.
\end{enumerate}

This paper will use $\mathbf{CC}$ to denote the information necessary
to capture the dependence of CrossCat on the data $\mathbf{X}$. This
includes the view concentration parameter $\alpha_D$, the
variable-specific hyper-parameters $\{\vec{\lambda}_d\}$, the view
partition $\vec{z}$, the view-specific concentration parameters
$\{\alpha_v\}$ and row partition $\{\vec{y}^v\}$, and the
category-specific parameters $\{\theta_c^d\}$ or sufficient statistics
$T_d(\vec{x}_{(\cdots,d)})$. This paper will also overload $ML_d, M_d,
V_d, L_d$, and $CRP$ to each represent both probability density
functions and stochastic simulators; the distinction should be clear
based on context. Given this notation, we have:
\hspace{-1.0in}%
\begin{flalign*}
P(\mathbf{CC}, \mathbf{X}) & = 
P(\mathbf{X}, \{\vec{\theta_c^d}\}, \{\vec{y}^v, \alpha_v\}, \{\vec{\lambda}_d\}, \vec{z}, \alpha_D) \\
& =  e^{-\alpha_D} \big( \prod_{d \in D} V_d(\vec{\lambda}_d) \big) \mathrm{CRP}(\vec{z} ; \alpha_D) \big( \prod_{v \in \vec{z}} e^{-\alpha_v} \mathrm{CRP}(\vec{y}^v ; \alpha_v) \big) \\
& \times 
\prod_{v \in \vec{z}} \prod_{c \in \vec{y}^v} \prod_{d \in \{i \mathrm{\ s.t.\ } z_i = v\}} 
\Big(
\begin{cases}
ML_d(T_d(\vec{x}_{(\cdot, d)}^c);  \vec{\lambda}_d) &\mathrm{if}\ \mathrm{u_d = 1} \\
M_d(\vec{\theta}_c^d ; \vec{\lambda}_d) \prod_{r \in c} L_d(x_{(r,d)} ; \vec{\theta_c^d}) &\mathrm{if}\ u_d = 0 \\
\end{cases}
\Big)
\end{flalign*}

\subsection{Hypothesis space and modeling capacity}

\begin{figure}[h]

\begin{center}
\includegraphics[width=6in]{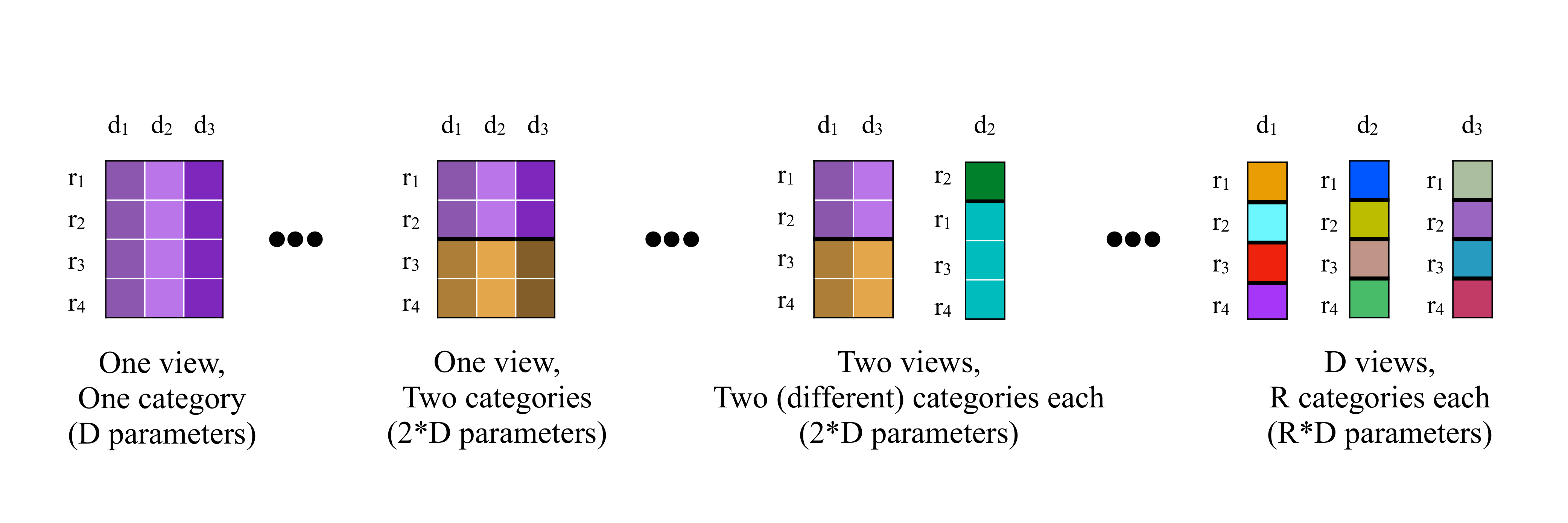}
\end{center}

\caption{ 
{\bf Model structures drawn from the space of all logically possible cross-categorizations of a 4 row, 3 column dataset.} In each structure, all data values (cells) that are governed by the same parametric model are shown in the same color. If two cells have different colors, they are modeled as conditionally independent given the model structure and hyper-parameters. In general, the space of all cross-categorizations contains a broad class of simple and complex data generators. See the main text for details.}
\label{fig:hypothesisSpace}
\end{figure}

The modeling assumptions encoded in CrossCat are designed to enable it to emulate a broad class of data generators. One way to assess this class is to study the full hypothesis space of CrossCat, that is, all logically possible cross-categorizations. Figure~\ref{fig:hypothesisSpace} illustrates the version of this space that is induced by a 4 row, 3 column dataset.  Each cross-categorization corresponds to a model structure --- a set of dependence and independence assumptions --- that is appropriate for some set of statistical situations. For example, conditioned on the hyper-parameters, the dependencies between variables and data values can be either dense or sparse. A group of dependencies will exhibit a unimodal joint distribution if they are modeled using only a single cluster. Strongly bimodal or multi-modal distributions as well as nearly unimodal distributions with some outliers are recovered by varying the number of clusters and their size. The prior favors stochastic relationships between groups of variables, but also supports (nearly) deterministic models; these correspond to structures with a large number of clusters that share low-entropy component models.

The CrossCat generative process favors hypotheses with multiple views
and multiple categories per view. A useful rule of thumb is to expect
$O(log(D))$ views with $O(log(R))$ categories each a priori. Asserting
that a dataset has several views and several categories per view
corresponds to asserting that the underlying data generator exhibits
several important statistical properties. The first is that the
dataset contains variables that arise from several distinct causal
processes, not just a single one. The second is that these processes
cannot be summarized by a single parametric model, and thus induce
non-Gaussian or multi-modal dependencies between the variables.

\subsection{Posterior Inference Algorithm} \label{inference}

\begin{figure}
\begin{center}
\includegraphics[width=6in]{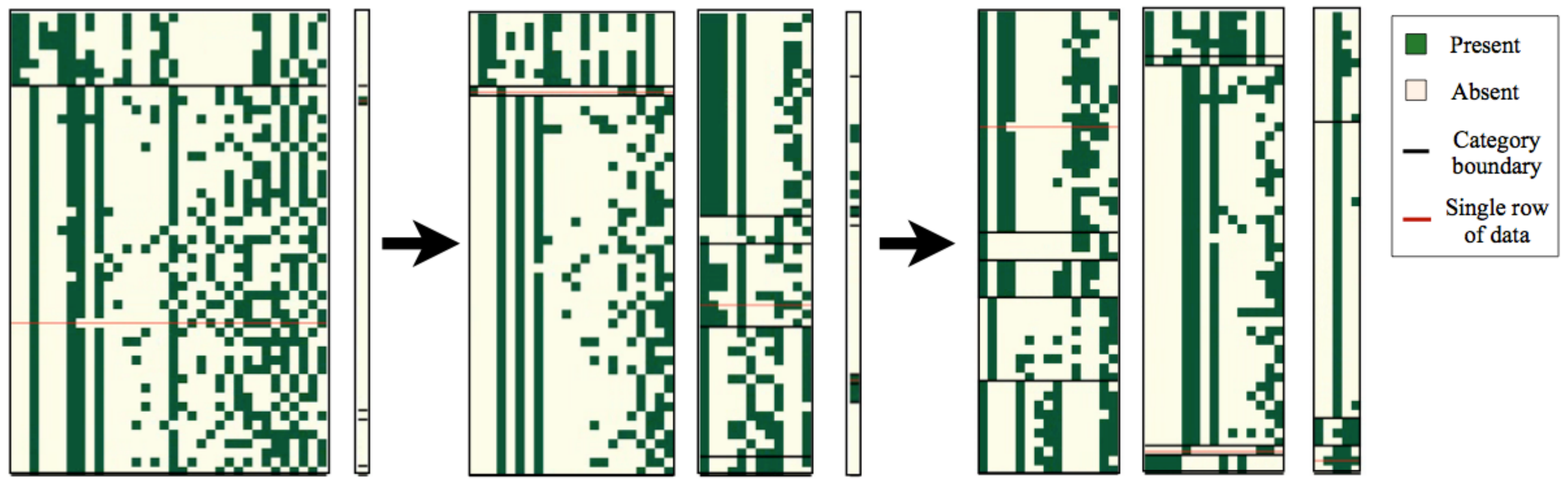}
\end{center}
\caption{{\bf Snapshots of the Markov chain for cross-categorization
    on a dataset of human object-feature judgments.} Each of the three states shows a
  particular cross-categorization that arose during a single Markov chain
  run, automatically rendered using the latent structure from
  cross-categorization to inform the layout. 
  Black horizontal lines separate categories within a view. The red horizontal line follows one row of the dataset.
  Taken from left to right,
  the three states span a typical run of roughly 100 iterations; the
  first is while the chain appears to be converging to a high
  probability region, while the last two illustrate variability within
  that region.}
\label{fig:mcmcmovie}
\end{figure}

Posterior inference is carried out by simulating an ergodic Markov chain that converges to the posterior \citep{Gilks99,Neal98}. The state of the Markov chain is a data structure storing the cross-categorization, sufficient statistics, and all uncollapsed parameters and hyper-parameters. Figure \ref{fig:mcmcmovie}
shows several sampled states from a typical run of the inference
scheme on the dataset from Figure~\ref{figure:crosscat-schematic}. 

The CrossCat inference Markov chain initializes a candidate state by
sampling it from the prior\footnote{Anecdotally, this initialization
  appears to yield the best inference performance overall. One
  explanation can be found by considering a representative subproblem
  of inference in CrossCat: performing inference in one of the inner
  CRP mixture models. A maximally dispersed initialization, with each
  of the $N$ rows in its own category, requires $O(N^2)$ time for its
  first Gibbs sweep. An initialization that places all rows in a
  single category requires $O(1)$ time for its first sweep but can
  spend many iterations stuck in or near the ``low resolution'' model
  encoded by this initial configuration.} The transition operator that
it iterates consists of an outer cycle of several kernels, each
performing cycle sweeps that apply other transition operators to each
segment of the latent state. The first is a cycle kernel for inference
over the outer CRP concentration parameter $\alpha$ and a cycle of
kernels over the inner CRP concentration parameters $\{ \alpha_v \}$
for each view. The second is a cycle of kernels for inference over the
hyper-parameters $\vec{\lambda}_d$ for each dimension. The third is a
kernel for inference over any uncollapsed parameters
$\vec{\theta}_c^d$. The fourth is a cycle over dimensions of an
inter-view auxiliary variable Gibbs kernel that shuffles dimensions
between views. The fifth is itself a cycle over views of cycles that
sweep a single-site Gibbs sampler over all the rows in the given
view. This chain corresponds to the default auxiliary variable Gibbs
sampler that the Venture probabilistic programming platform 
\citep{mansinghka2014venture} produces when given the CrossCat model 
written as a probabilistic program.

More formally, the Markov chain used for inference is a cycle over the
following kernels:

\begin{enumerate}
\item {\bf Concentration hyper-parameter inference: updating $\alpha_D$ and each element of $\{\alpha_v\}$.} Sample $\alpha_D$ and all the $\alpha_v$s for each view via a discretized Gibbs approximation to the posterior,\ $\alpha_D \sim P(\alpha_D|\vec{z})$ and $\alpha_v \sim P(\alpha_v|\vec{y^v})$. For each $\alpha$, this involves scoring the CRP marginal likelihood at a fixed number of grid points --- typically $\sim100$ --- and then re-normalizing and sampling from the resulting discrete approximation.

\item {\bf Component model hyper-parameter inference: updating the elements of $\{\vec{\lambda}_d\}$.} For each dimension, for each hyper-parameter, discretize the support of the hyper-prior and numerically sample from an approximate hyper-posterior distribution. That is, implement an appropriately-binned discrete approximation to a Gibbs sampler for $\vec{\lambda}_d \sim P(\vec{\lambda}_d \given \vec{x_d}, \vec{y^{z_d}})$ (i.e. we condition on the vertical slice of the input table described by the hyper-parameters, and the associated latent variables). For conjugate component models, the probabilities depend only on the sufficient statistics needed to evaluate this posterior. Each hyper-parameter adjustment requires an operation linear in the number of categories, since the scores for each category (i.e.\ the marginal probabilities) must be recalculated, after each category's statistics are updated. Thus each application of this kernel takes time proportional to the number of dimensions times the maximum number of categories in any view.

\item {\bf Category inference: updating the elements of
    $\{\vec{y^v}\}$ via Gibbs with auxiliary variables.} For each
  entity in each view, this transition operator samples a new category
  assignment from its conditional posterior. A variant of Algorithm 8 from \citep{Neal98} (with $m$=1) is used to handle uncollapsed dimensions.

  The category inference transition operator will sample $y^v_r$, the
  categorization for row $r$ in view $v$, according to its conditioned
  distribution given the other category assignments $\vec{y^v}_{-r}$,
  parameters $\{\vec{\theta_c^d}\}$ and auxiliary parameters. If $u_d
  = 0\ \forall d\ s.t.\ z_d = v$, i.e. there are no uncollapsed
  dimensions in this view, then this reduces to the usual collapsed
  Gibbs sampler applied to the subset of data within the
  view. Otherwise, let $\{\vec{\phi}^d\}$ denote auxiliary parameters
  for each uncollapsed dimension $d$ (where $u_d = 1$) of the same
  form as $\vec{\theta_c^d}$. Before each transition, these
  parameters are chosen as follows:
\begin{align*}
\vec{\phi}^d & \sim \begin{cases}
\delta_{\vec{\theta_{y^v_r}^d}} & \mathrm{if}\ y^v_r = y_j^v \iff r = j \\
 M_d(\vec{\lambda}_d) & o.w.\ (y_r^v \in \vec{y}_{-r}^v) 
\end{cases}
\end{align*}
In this section, $c^+$ will denote the category associated with the
auxiliary variable. If $y_r^v \in \vec{y}^v_{-r}$, then $c^+ =
\mathrm{max}(\vec{y}_v^{-r}) + 1$, i.e. a wholly new category will be
created, and by sampling $\vec{\phi}^d$ this category will have
newly sampled parameters. Otherwise, $c^+ = y_r^v$, i.e. row $r$ was a
singleton, so its previous category assignment and parameters will
be reused.

Given the auxiliary variables, we can derive the target density of the
transition operator by expanding the joint probability density:
\begin{align*}
y_r^v & 
\sim P(y_r^v \given \vec{y}^v_{-r}, \{\vec{\lambda}_d, \{x_{(\cdot, d)}\} \mid d\ s.t.\ z_d = v\}, \{\{\vec{\theta_c^d} \mid c \in \vec{y}^v_{-r}\} \mid d\ s.t.\ z_d = v\ \mathrm{and}\ u_d = 1\}, \{\vec{\phi}^d\}) \\
 & \propto \mathrm{CRP}(y_r^v; \vec{y}^v_{-r}, \alpha_v) \prod_{d \in \{i \mathrm{\ s.t.\ } z_i = v\}} \Big(
\begin{cases}
ML_d(T_d(\vec{x}_{(\cdot, d)}^c), \vec{\lambda}_d) &\mathrm{if}\ u_d = 0 \\
M_d(\vec{\theta}_c^d ; \vec{\lambda}_d) \prod_{r \in c} L_d(x_{(r,d)} ; \vec{\theta_c^d}) &\mathrm{if}\ u_d = 1\ \mathrm{and}\ y_r^v \in \vec{y}^v_{-r}\\
M_d(\vec{\phi}_c^d ; \vec{\lambda}_d) \prod_{r \in c} L_d(x_{(r,d)} ; \vec{\phi_c^d}) &\mathrm{if}\ u_d = 1\ \mathrm{and}\ y_r^v = c^+ \notin \vec{y}^v_{-r}
\end{cases}
\Big)
\end{align*}

The probabilities this transition operator needs can be obtained by
iterating over possible values for $y_r^v$, calculating their joint
densities, and re-normalizing numerically. These operations can be
implemented efficiently by maintaining and incrementally modifying a
representation of $\mathbf{CC}$, updating sufficient statistics and a
joint probability accumulator after each change
\citep{mansinghka2007}. The complexity of resampling $y_r^v$ for all
rows $r$ and views $v$ is $O(VRCD)$, where $V$ is the number of views,
$R$ the number of rows, $C$ the maximum number of categories in any
view, and $D$ is the number of dimensions.

\item {\bf Inter-view inference: updating the elements of $\vec{z}$
    via Gibbs with auxiliary variables.} For each dimension $d$, this
  transition operator samples a new view assignment $z_d$ from its
  conditional posterior. As with the category inference kernel, this can be
  viewed as a variant of Algorithm 8 from \citep{Neal98} (with $m=1$),
  applied to the ``outer'' Dirichlet process mixture model in
  CrossCat. This mixture has uncollapsed, non-conjugate component
  models that are themselves Dirichlet process mixtures.

  Let $v^+$ be the index of the new view. The auxiliary variables are
  $\alpha_{v^+}$, $\vec{y}^{v^+}$ and $\{\theta_c^d \mid c \in
  \vec{y}^{v^+}\}$ (if $u_d = 1$). If $z_d \in \vec{z}^{-d}$, then $v^+
    = \mathrm{max}(\vec{z}) + 1$, and the auxiliary variables are
    sampled from their priors. Otherwise, $v^+ = z_d$, and the
    auxiliary variables are deterministically set to the values
    associated with $z_d$. Given values for these variables, the
    conditional distribution for $z_d$ can be derived as follows:

\begin{align*}
z_d & \sim P(z_d \given \alpha_D, \vec{\lambda}_d, \vec{z}^{-d}, \alpha_{v^+}, \{\vec{y}^v\},
\{\{\theta_c^d \mid c \in \vec{y}^{z_j}\} \mid j \in D\}, \mathbf{X})\\
    & \propto \mathrm{CRP}(z_d; \vec{z^{-d}}, \alpha_D) \prod_{c \in \vec{y}^{z_d}} 
\Big(
\begin{cases}
ML_d(T_d(\vec{x}_{(\cdot, d)}^c), \vec{\lambda}_d) &\mathrm{if}\ \mathrm{u_d = 1} \\
M_d(\vec{\theta}_c^d ; \vec{\lambda}_d) \prod_{r \in c} L_d(x_{(r,d)} ; \vec{\theta_c^d}) &\mathrm{if}\ u_d = 0 \\
\end{cases}
\Big)
\end{align*}

This transition operator shuffles individual columns between views,
weighing their compatibility with each view by multiplying likelihoods
for each category. A full sweep thus has time complexity
$O(DVCR)$. Note that if a given variable is a poor fit for its current
view, its hyper-parameters and parameters will be driven to reduce the
dependence of the likelihood for that variable on its clustering. This
makes it more likely for row categorizations proposed from the prior
to be accepted.

Inference over the elements of $\vec{z}$ can also be done via a
mixture of a Metropolis-Hastings birth-death kernel to create new
views with a standard Gibbs kernel to reassign dimensions among
pre-existing views. In our experience, both transition operators yield
comparable results on real-world data; the Gibbs auxiliary variable
kernel is presented here for simplicity.

\item {\bf Component model parameter inference: updating
    $\{\vec{\theta_c^d} \mid u_d = 1\}$.} Each dimension or variable
  whose component models are uncollapsed must be equipped with a
  suitable ergodic transition operator $T$ that converges to the local
  parameter posterior $P(\vec{\theta_c^d} | \vec{x_{(\cdot, d)}^c},
  \vec{\lambda}_d)$. Exact Gibbs sampling is often possible when $L_d$
  and $M_d$ are conjugate.
\end{enumerate}

CrossCat's scalability can be assessed by multiplying an estimate of
how long each transition takes with an estimate of how many
transitions are needed to get good results. The experiments in this
paper use $\sim$10-100 independent samples. Each sample was based on
runs of the inference Markov chain with $\sim$100-1,000
transitions. Taking these numbers as rough constants, scalability is
governed by the asymptotic orders of growth. Let $R$ be the number of
rows, $D$ the number of dimensions, $V$ the maximum number of views
and $C$ the maximum number of categories. The memory needed to store
the latent state is the sum of the memory needed to store the $D$
hyper-parameters and view assignments, the $VC$ parameters/sufficient
statistics, and the $VR$ category assignments, or $O(D + VC +
VR)$. Assuming a fully dense data matrix, the loops in the transition
operator described above scale as $O(DC + RDVC + RDVC + DC) =
O(RDVC)$, with the $RD$ terms scaling down following the data density.

This paper shows results from both open-source and commercial
implementations on datasets of up to $\sim$10 million cells\footnote{A
  variation on CrossCat was the basis of Veritable, a general-purpose
  machine learning system built by Navia Systems/Prior Knowledge
  Inc. This implementation became a part of Salesforce.com's
  predictive analytics infrastructure. At Navia, CrossCat was applied
  to proprietary datasets from domains such as operations management
  for retail, clinical virology, and quantitative finance.}.  Because
this algorithm is asymptotically linear in runtime with low memory
requirements, a number of performance engineering and distributed
techniques can be applied to reach larger scales at low
latencies. Performance engineering details are beyond the scope of
this paper.

\subsection{Exploration and Prediction Using Posterior Samples}

Each approximate posterior sample provides an estimate of the full
joint distribution of the data. It also contains a candidate latent
structure that characterizes the dependencies between variables and
provides an independent clustering of the rows with respect to each
group of dependent variables. This section gives examples of
exploratory and predictive analysis problems that can be solved by
using these samples. Prediction is based on calculating or sampling
from the conditional densities implied by each sample and then either
averaging or resampling from the results. Exploratory queries
typically involve Monte Carlo estimation of posterior probabilities
that assess structural properties of the latent variables posited by
CrossCat and the dependencies they imply. Examples include obtaining a
global map of the pairwise dependencies between variables, selecting
those variables that are probably predictive of some target, and
identifying rows that are similar in light of some variables of
interest.

\subsubsection{Prediction}

Recall that $\mathbf{CC}$ represents a model for the joint distribution over the variables along with sufficient statistics, parameters, a partition of variables into views, and categorizations of the rows in the data $\mathbf{X}$. Variables representing the latent structure associated with a particular posterior sample $\hat{CC}_s$ will all be indexed by $s$, e.g. $z_d^s$. Also let $Y_v^+$ represent the category assignment of a new row in view $v$, and let $\{t_i\}$ and $\{g_j\}$ be the sets of target variables and given variables in a given predictive query.

To generate predictions by sampling from the conditional density on targets given the data, we must simulate
$$
\{ \hat{x}_{t_i} \} \sim p( \{ X_{t_i} \} | \{ X_{g_i} = x_{g_i} \}, \mathbf{X})
$$
Given a set of models, this can be done in two steps. First, from each model, sample a categorization from each view conditioned on the values of the given variables. Second, sample values for each target variable by simulating from the target variable's component model for the sampled category:

\begin{eqnarray*}
\hat{CC}_s & \sim & p( \mathbf{CC} | \mathbf{X} ) \\
c_v^s & \sim & p( Y_v^+ | \{ X_{g_j} = x_{g_j} | z_{g_j}^s = v \} ) \\
\hat{x}_{t_i}^s & \sim & p( X_{t_i} | c_{z_{t_i}}^s ) = \int 
L(x_{t_i}; \vec{\theta}_{c_v^s}^{t_i}) M(\vec{\theta}_{c_v^s}^{t_i} ; \vec{\lambda_{t_i}}) d\vec{\theta}  
\end{eqnarray*}

The category kernel from the MCMC inference algorithm can be re-used to sample from $c_v^s$. Also, sampling from $\hat{x}_{t_i}^s$ can be done directly given the sufficient statistics for data types whose likelihood models and parameter priors are conjugate. In other cases, either $\vec{\theta}$ will be represented as part of $\hat{CC}_s$ or sampled on demand.

The same latent variables are also useful for evaluating the conditional density for a desired set of predictions:

\begin{eqnarray*}
p( \{ X_{t_i} = x_{t_i} \} | \{ X_{g_j} = x_{g_j} \}, \mathbf{X}) & \approx & \frac{1}{N}\sum_{s} p( \{ X_{t_i} = x_{t_i} \} | \{ X_{g_j} = x_{g_j} \}, \mathbf{CC} = \hat{CC}_s ) \\
& = & \frac{1}{N}\sum_{s} \prod_{v \in \vec{z}^s} \sum_{c} p( \{ X_{t_i} = x_{t_i}  | z_{g_j}^s = v \} | Y_v^+ = c ) p( Y_v^+ = c |  \{ X_{g_j} = x_{g_j} | z_{g_j}^s = v \} )
\end{eqnarray*}

Many problems of prediction can be reduced to sampling from and/or
calculating conditional densities. Examples include classification,
regression and imputation. Each can be implemented by forming
estimates $ \{X_{t_i}^*\}$ of the target variables. By default, the
implementation from this paper implementation uses the mean of the
predictive to impute continuous values. This is equivalent to choosing
the value that minimizes the expected square loss under the empirical
distribution induced by a set of predictive samples. For discrete
values, the implementation uses the most probable value, equivalent to
minimizing 0-1 loss, and calculates it by directly evaluating the
conditional density of each possible value. This approach to
prediction can also handle nonlinear and/or stochastic relationships
within the set of target variables $\{ X_{t_i} \}$ and between the
given variables $\{ X_{g_i} \}$ and the targets. It is easy to
implement in terms of the same sampling and probability calculation
kernels that are necessary for inference.

This formulation of prediction scales linearly in the number of
variables, categories, and view. It is also sub-linear in the number
of variables when dependencies are sparse, and parallelizable over the
views, the posterior samples, and the generated samples from the
conditional density. Future work will explore the space of tradeoffs
between accuracy, latency and throughput that can be achieved using
this basic design.

\subsubsection{Detecting dependencies between variables}

To detect dependencies between groups of variables, it is natural to
use a Monte Carlo estimate of the marginal posterior probability that
a set of variables $\{ q_i \}$ share the same posterior view. Using
$s$ as a superscript to select values from a specific sample, we have:
\begin{eqnarray*}
Pr[ z_{q_0} = z_{q_1} = \cdots = z_{q_k} | \mathbf{X} ] & \approx & \frac{1}{N} \sum_s Pr[ z_{q_0}^s = z_{q_1}^s = \cdots = z_{q_k}^s | \hat{CC}_s ]  \\
& = & \frac{ \#( \{ s | z_{q_0}^s = z_{q_1}^s = \cdots = z_{q_k}^s  \} ) }{N}
\end{eqnarray*}
These probabilities also characterize the marginal dependencies and
independencies that are explicitly represented by CrossCat. For
example, pairwise co-assignment in $\vec{z}$ determines\footnote{This
  paper defines independence in terms of the generative process and
  latent variables. Two variables in different views are explicitly
  independent, but two variables in the same view are coupled through
  the latent cluster assignment. This is clear if there are multiple
  clusters. Even if there is just one cluster, if $\alpha_v$ remains
  nonzero as $N$ goes to infinity, then eventually there will be more
  than one cluster. A predictive definition of independence in terms
  of nonzero mutual information will differ in some cases; a
  comparison between these candidate measures is beyond the scope of
  this paper..} pairwise marginal independence under the generative
model:
$$
X_{q_i} \bigCI X_{q_j} \iff z_{q_i} \neq z_{q_k}
$$
The results in this paper often include the ``z-matrix'' of marginal
dependence probabilities $\mathbf{Z} = [ Z_{(i,j)} ]$, where
$Z_{(i,j)} = 1 - Pr[ X_i \bigCI X_j | \mathbf{X} ]$. This measure is
used primarily for simplicity; other measures of the presence or
strength of predictive relationships are possible.

\subsubsection{Estimating similarity between rows}

Exploratory analyses often make use of ``similarity'' functions
defined over pairs of rows. One useful measure of similarity is given
by the probability that two pieces of data were generated from the
same statistical model \citep{tenenbaumgriffiths01,
  ghahramani2006bayesian}. CrossCat naturally induces a
context-sensitive similarity measure between rows that has this form:
the probability that two items come from the same category in some
context. Here, contexts are defined by target variables, and comprise
the set of views in which that variable participates (weighted by
their probability). This probability is straightforward to estimate
given a collection of samples:

$$
1 - Pr[x_{(r, c)} \bigCI x_{(r', c)} | \mathbf{X}, \vec{\lambda}_c] \approx \frac{\#( \{ s | y^{z_c^s}_{(s,r)} = y^{z_c^s}_{(s,r')} \} )}{N}
$$

\noindent This measure relies on CrossCat's detection of marginal
dependencies to determine which variables are relevant in any given
context. The component models largely determine how differences in each
variable in that view will be weighted when calculating similarity.

\subsection{Assessing Inference Quality}

A central concern is that the single-site Gibbs sampler used for
inference might not produce high-quality models or stable posterior
estimates within practical running times. For example, the CrossCat
inference algorithm might rapidly converge to a local minimum in which
all proposals to create new views are rejected. In this case, even
though the Gibbs sampler will appear to have converged, the models it
produces could yield poor inference quality.

This section reports four experiments that illustrate key algorithmic
and statistical properties of CrossCat. The first experiment gives a
rough sense of inference efficiency by comparing the energies of
ground truth states to the energies of states sampled from CrossCat on
data generated by the model. The second experiment assesses the
convergence rate and the reliability of estimates of posterior
expectations on a real-world dataset. The third experiment explores
CrossCat's resistance to under-fitting and over-fitting by running
inference on datasets of Gaussian noise. The fourth experiment
assesses CrossCat's predictive accuracy in a setting with a large
number of distractors and a small number of signal variables. It shows
that CrossCat yields favorable accuracy compared to several baseline
methods.

\begin{figure}[ht]
\centering
\includegraphics[scale=.3]{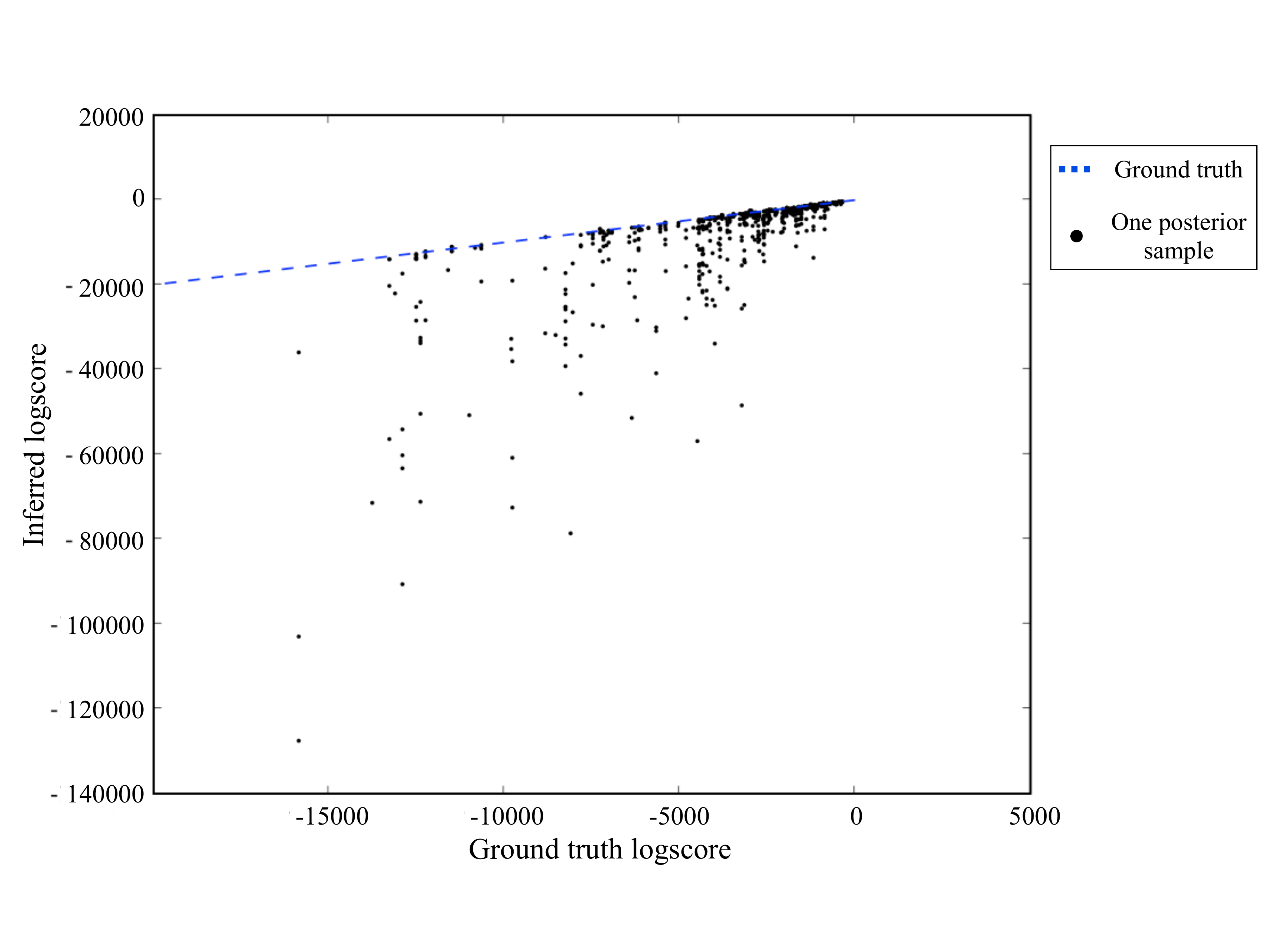}
\caption{{\bf The joint density of the latent cross-categorization and
    training data for $\sim$1,000 samples from CrossCat's inference
    algorithm, compared to ground truth.} Each point corresponds to a
  sample drawn from an approximation of the CrossCat posterior
  distribution after 200 iterations on data from a randomly chosen
  CrossCat model. Table sizes range from 10x10 to 512x512. Points on
  the blue line correspond to samples with the same joint density as
  the ground truth state. Points lying above the line correspond to
  models that most likely underestimate the entropy of the underlying
  generator, i.e. they have over-fit the data. CrossCat rarely produces
  such samples. Some points lie significantly below the line,
  overestimating the entropy of the generator. These do not
  necessarily correspond to ``under-fit'' models, as the true posterior
  will be broad (and may also induce broad predictions) when data is
  scarce.}
\label{fig:synthetic_mix}
\end{figure}

\begin{figure}[t]
\centering
(a) \includegraphics[scale=0.3]{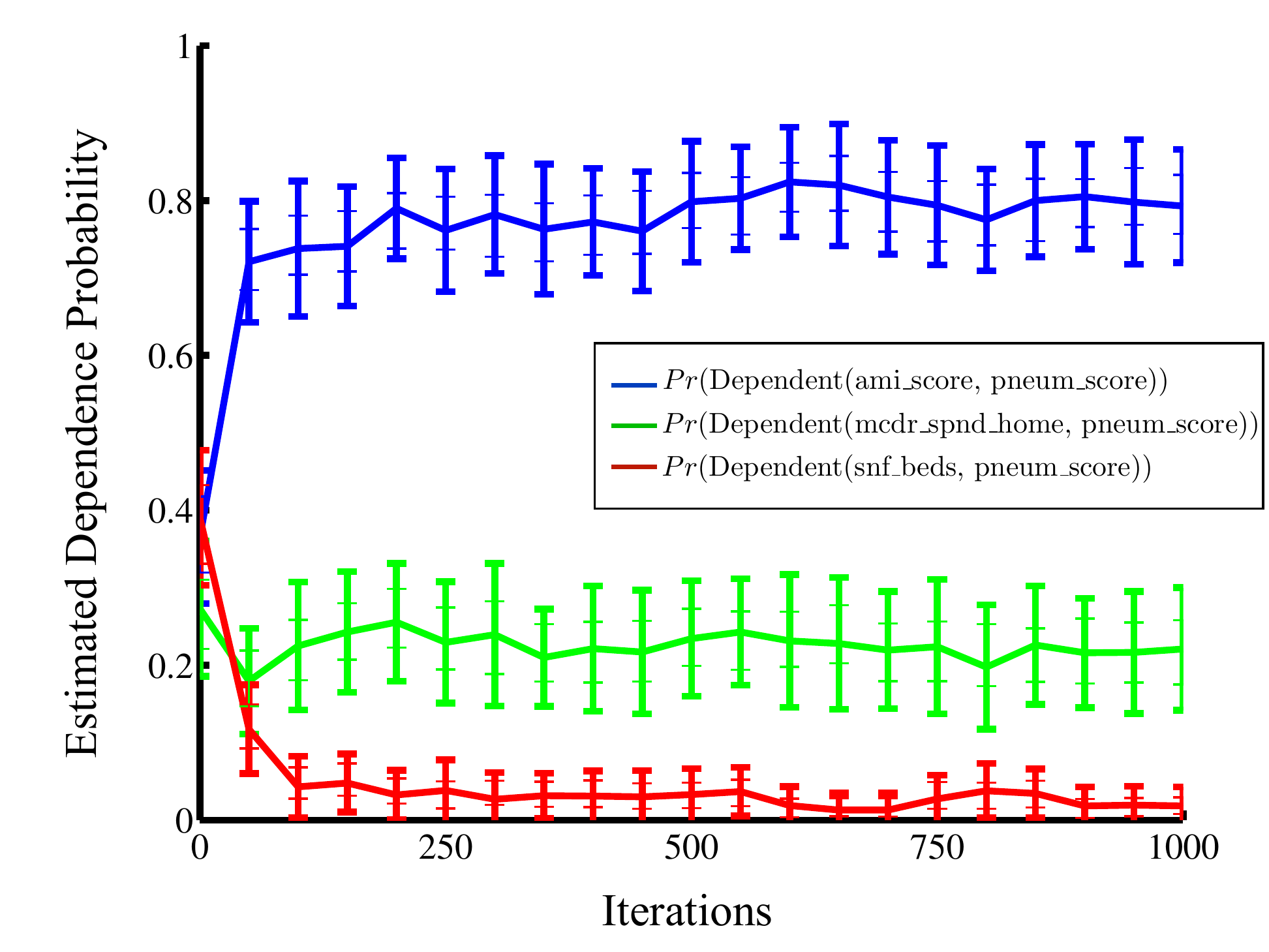} 
(b) \includegraphics[scale=0.25]{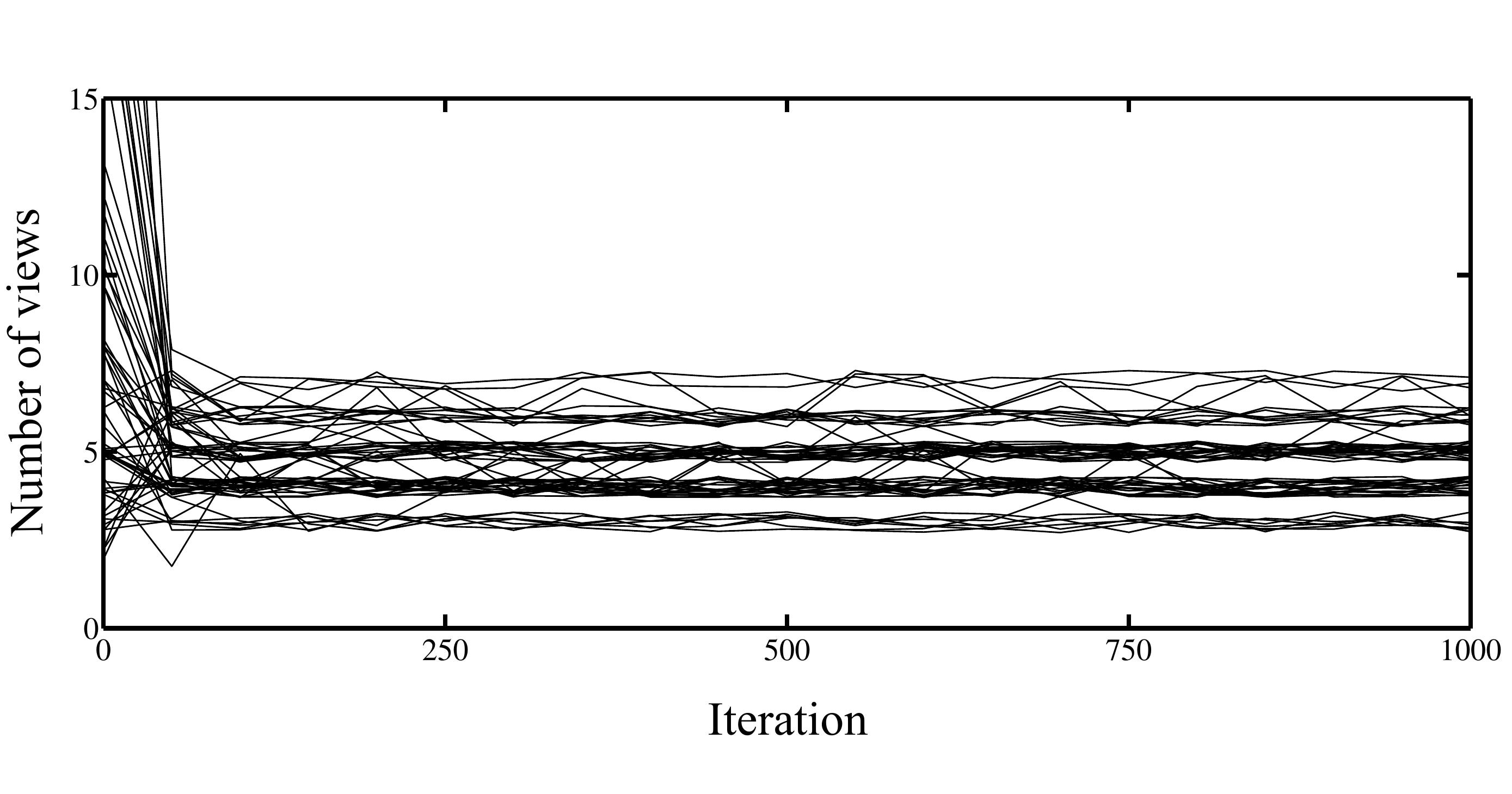}
\caption{{\bf A quantitative assessment of the convergence rate of CrossCat inference and the stability of posterior estimates on a real-world health economics dataset.} (a) shows the evolution of simple Monte Carlo estimates of the probability of dependence of three pairs of variables, made from independent chains initialized from the prior, as a function of the number of iterations of inference. Thick error bars show the standard deviation of estimates across 50 repetitions, each with 20 samples; thin lines show the standard deviation of estimates from 40 samples. 
Estimates stabilize after $\sim$100 iterations. (b) shows the number of views for 50 of the same Markov chain runs. After $\sim$100 iterations, states with 4, 5 or 6 views dominate the sample, and chains still can switch into and out of this region after 700 iterations.}
\label{fig:dha_convergence}
\end{figure}

The next experiment assesses the stability and efficiency of CrossCat
inference on real-world data. Figure~\ref{fig:dha_convergence}a shows
the evolution of Monte Carlo dependence probability estimates as a
function of the number of Markov chain
iterations. Figure~\ref{fig:dha_convergence}b shows traces of the
number of views for each chain in the same set of runs. $\sim$100
iterations appears sufficient for initializations to be forgotten,
regardless of the number of views sampled from the CrossCat prior. At
this point, Monte Carlo estimates appear to stabilize, and the
majority of states ($\sim$40 of 50 total) appear to have 4, 5 or 6
views. This stability is not simply due to a local minimum: after 700
iterations, transitions that create or destroy views are still being
accepted. However, the frequency of these transitions does
decrease. It thus seems likely that the standard MCMC approach of
averaging over a single long chain run might require significantly
more computation than parallel chains. This behavior is typical for
applications to real-world data. We typically use 10-100 chains, each
run for 100-1,000 iterations, and have consistently obtained stable
estimates.

The convergence measures from \citep{geweke92} are also included for
comparison, specifically the numerical standard error (NSE) and
relative numerical efficiency (RNE) for the view CRP parameter
$\alpha$ to assess autocorrelations \citep{lesage99}. NSE values near
0 and RNE values near 1 indicate approximately independent
draws. These values were computed using a $0\%$, $4\%$, $8\%$, and
$15\%$ autocorrelation taper. NSE values were near zero and did not
differ markedly: $.023$, $.021$, $.018$, and $.018$. Similarly, RSE
values were near 1 and did not differ markedly: $1$, $1.23$, $1.66$,
and $1.54$. These results suggest that there is acceptably low
autocorrelation in the sampled values of the hyper-parameters.

The reliability of CrossCat reflects simple but important differences
between the way single-site Gibbs sampling is used here and standard
MCMC practice in machine learning. First, CrossCat uses independent
samples from parallel chains, each initialized with an independent
sample from the CrossCat prior. In contrast, typical MCMC schemes from
nonparametric Bayesian statistics use dependent samples obtained by
thinning a single long chain that was deterministically
initialized. For example, Gibbs samplers for Dirichlet process
mixtures are often initialized to a state with a single cluster; this
corresponds to a single-view single-category state for
CrossCat. Second, CrossCat performs inference over hyper-parameters
that control the expected predictability of each dimension, as well as
the concentration parameters of all Dirichlet processes. Many machine
learning applications of nonparametric Bayes do not include inference
over these hyper-parameters; instead, they are set via cross-validation
or other heuristics.

There are mechanisms by which these differences could potentially
explain the surprising reliability and speed of CrossCat inference as
compared to typical Gibbs samplers. Recall that the regeneration time
of a Markov chain started at its equilibrium distribution is the
(random) amount of time it needs to ``forget'' its current state and
arrive at an independent sample. For CrossCat, this regeneration time
appears to be substantially longer than convergence time from the
prior. States from the prior are unlikely to have high energy or be
near high energy regions, unlike states drawn from the
posterior. Second, hyper-parameter inference --- especially those
controlling the expected noise in the component models, not just the
Dirichlet process concentrations --- provides a simple mechanism that
helps the sampler exit local minima. Consider a dimension that is
poorly explained by the categorization in its current
view. Conditioned on such a categorization, the posterior on the
hyper-parameter will favor increasing the expected noisiness of the
clusters, to better accommodate the data. Once the hyper-parameter
enters this regime, the model becomes less sensitive to the specific
clustering used to explain this dimension. This therefore also
increases the probability that the dimension will be reassigned to any
other pre-existing view. It also increases the acceptance probability
for proposals that create a new view with a random
categorization. Once a satisfactory categorization is found, however,
the Bayesian Occam's Razor favors reducing the expected entropy of the
clusters. Similar dynamics were described in
\citep{mansinghka2013approximate}; a detailed study is beyond the
scope of this paper.

The third simulation, shown in Figure~\ref{fig:needle}, illustrates
CrossCat's behavior on datasets with low-dimensional signals amidst
high-dimensional random noise. In each case, CrossCat rapidly and
confidently detects the independence between the ``distractor''
dimensions, i.e. it does not over-fit. Also, when the signal is strong
or there are few distractors, CrossCat confidently detects the true
predictive relationships. As the signals become weaker, CrossCat's
confidence decreases, and variation increases. These examples
qualitatively support the use of CrossCat's estimates of dependence
probabilities as indicators of the presence or absence of predictive
relationships. A quantitative characterization of CrossCat's
sensitivity and specificity, as a function of both sample size and
strength of dependence, is beyond the scope of this paper.

\begin{figure}
\hspace{-0.25in}
\begin{minipage}{0.25\textwidth}
  1.0 - Pr[$X_i \bigCI X_k | \mathbf{X} ]$ \\ for strong signal, \\ many distractors \\ ($\rho = 0.7$, $D = 20$) \\
\\\\
Moderate signal, \\ few distractors \\ ($\rho = 0.5$, $D = 8$) \\
\\\\\\
Moderate signal, \\ many distractors \\ ($\rho = 0.5$, $D = 20$) \\
\\\\\\
Weak signal, \\ few distractors \\ ($\rho = 0.25$, $D = 8$) \\
\end{minipage}
\begin{minipage}{0.65\textwidth}
\includegraphics[width=5in]{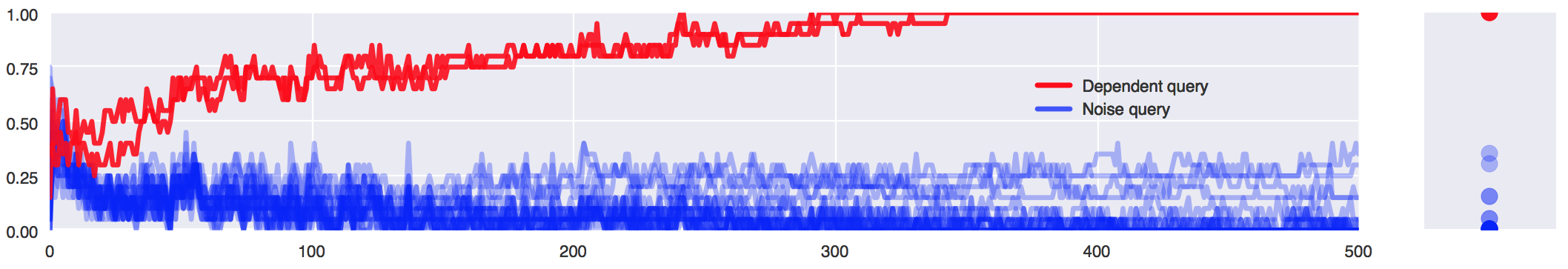}\\
\\
\includegraphics[width=5in]{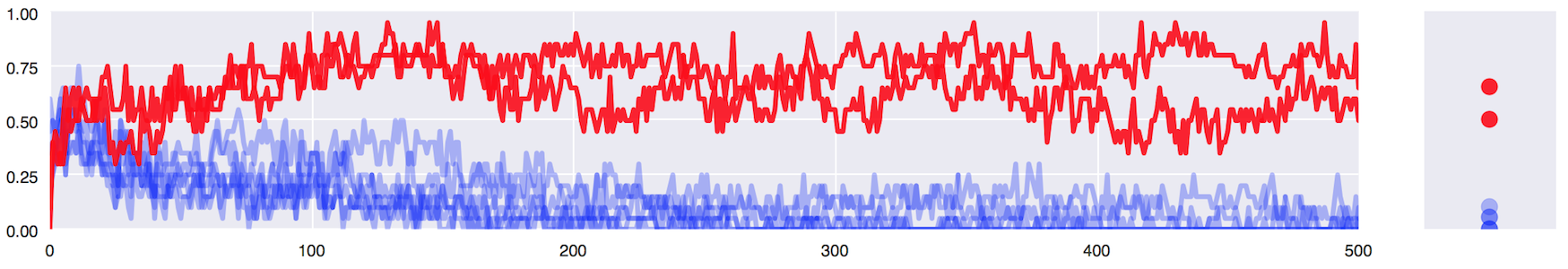}\\
\\
\includegraphics[width=5in]{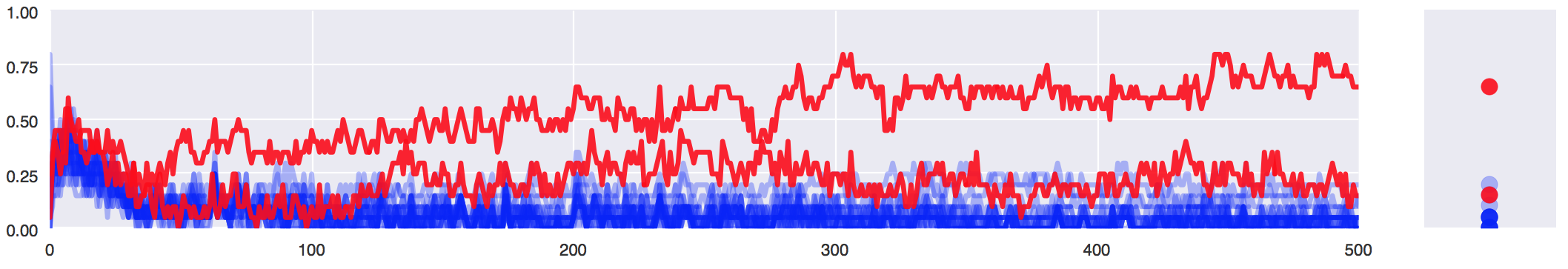}\\
\\
\includegraphics[width=5in]{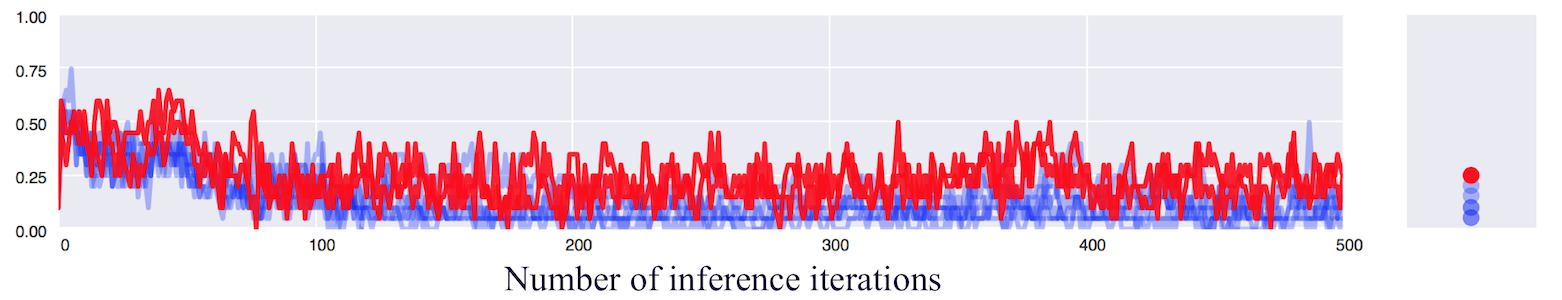}
\end{minipage}

\caption{{\bf Detected dependencies given two correlated signal
    variables and multiple independent distractors.} This experiment
  illustrates CrossCat's sensitivity and specificity to pairwise
  relationships on multivariate Gaussian datasets with 100 rows. In
  each dataset, two pairs of variables have nonzero correlation
  $\rho$. The remaining $D - 4$ dimensions are uncorrelated
  distractors. Each row shows the inferred dependencies between 20
  randomly sampled pairs of distractors (blue) and the two pairs of
  signal variables (red). See main text for further discussion.}
\label{fig:needle}
\end{figure}

Many data analysis problems require sifting through a large pool of
candidate variables in settings where only a small fraction are
relevant for any given prediction. The fourth experiment, shown in
Figure \ref{fig:needle}, illustrates CrossCat's behavior in this
setting. The test datasets contain 10 ``signal'' dimensions generated
from a 5-component mixture model, plus 10-1,000 ``distractor''
dimensions generated by an independent 3-component mixture that
clusters the data differently. As the number of distractors increases,
the likelihood becomes dominated by the distractors. The experiment
compares imputation accuracy for several methods --- CrossCat; mixture
modeling; column-wise averaging; imputation by randomly chosen values;
and a popular model-free imputation technique
\citep{Hastie99imputingmissing} --- on problems with varying numbers
of distractors. CrossCat remains accurate when the number of
distractors is 100x larger than the number of signal variables. As
expected, mixtures are effective in low dimensions, but inaccurate in
high dimensions. When the number of distractors equals the number of
signal variables, the mixture posterior grows bimodal, including one
mode that treats the signal variables as noise. This mode dominates
when the number of distractors increases further.

\begin{figure}
\centering
\includegraphics[width=6in]{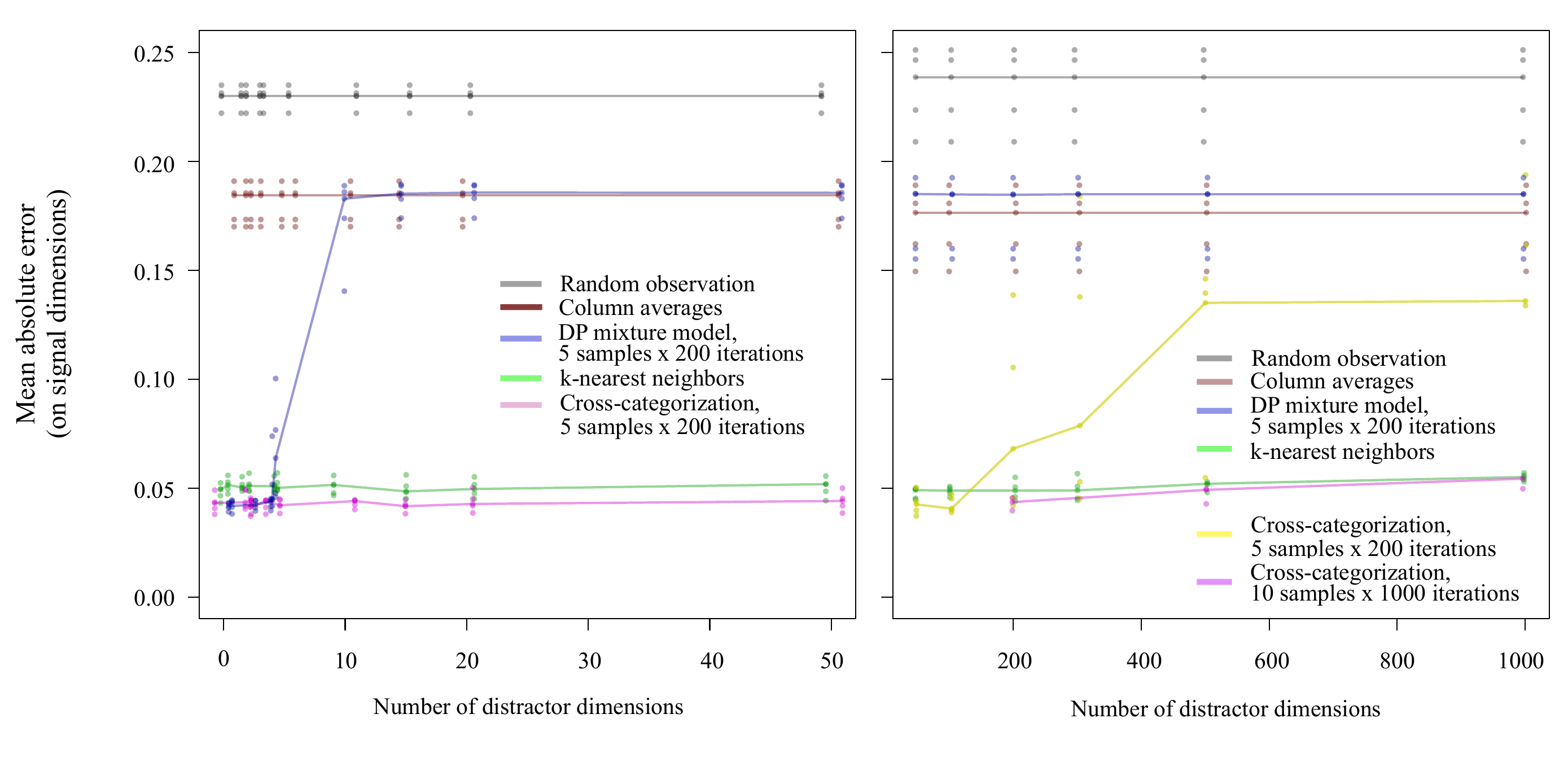}
\caption{ {\bf Predictive accuracy for low-dimensional signals
    embedded in high-dimensional noise.} The data generator contains
  10 ``signal'' dimensions described by a 5-cluster model to which
  distractor dimensions described by an independent 3-cluster model
  have been appended. The left plot shows imputation accuracies for up
  to 50 distractor dimensions; the right shows accuracies for 50-1,000
  distractors. CrossCat is compared to mixture models as well as
  multiple non-probabilistic baselines (column-wise averaging,
  imputing via a random value, and a state-of-the-art extension to
  k-nearest-neighbors). The accuracy of mixture modeling drops when
  the number of distractors D becomes comparable to the number of
  signal variables S, i.e. when D$\ \approx\ $S. When D $>$ S, the
  distractors get modeled instead of the signal. In contrast, CrossCat
  remains accurate when the number of distractors is 100 times larger
  than the number of signal variables. See main text for additional
  discussion.}
\label{fig:needle}
\end{figure}

\FloatBarrier

\section{Empirical Results on Real-World Datasets}

This section describes the results from CrossCat-based analyses of
several datasets. Examples are drawn from multiple fields, including
health economics, pattern recognition, political science, and
econometrics. These examples involve both exploratory analysis and
predictive modeling. The primary aim is to illustrate CrossCat and
assess its efficacy on real-world problems. A secondary aim is to
verify that CrossCat produces useful results on data generating
processes with diverse statistical characteristics. A third aim is to
compare CrossCat with standard generative, discriminative, and
model-free methods.

\subsection{Dartmouth Atlas of Health Care}

The Dartmouth Atlas of Health Care \citep{Dha} is one output from a
long-running effort to understand the efficiency and effectiveness of
the US health care system. The overall dataset covers $\sim$4300
hospitals that can be aggregated into $\sim$300 hospital reporting
regions. The extract analyzed here contains 74 variables that
collectively describe a hospital's capacity, quality of care, and cost
structure. These variables contain information about multiple
functional units of a hospital, such as the intensive care unit (ICU),
its surgery department, and any hospice services it offers. For
several of these units, the amount each hospital bills to a federal
program called Medicare is also available. The continuous variables in
this dataset range over multiple orders of magnitude. Specific
examples include counts of patients, counts of beds, dollar amounts,
percentages that are ratios of counts in the dataset, and numerical
aggregates from survey instruments that assess quality of care.

Due to its broad coverage of hospitals and their key characteristics,
this dataset illustrates some of the opportunities and challenges
described by the \citet{napmassive}. For example, given the range of
cost variables and quality surveys it contains, this data could be
used to study the relationship between cost and quality of care. The
credibility of any resulting inferences would rest partly on the
comprehensiveness of the dataset in both rows (hospitals) and columns
(variables). However, it can be difficult to establish the absence of
meaningful predictive relationships in high-dimensional data on purely
empirical grounds. Many possible sets of predictors and forms of
relationships need to be considered and rejected, without sacrificing
either sensitivity or specificity. If the dataset had fewer variables,
a negative finding would be easier to establish, both statistically
and computationally, as there are fewer possibilities to
consider. However, such a negative finding would be less convincing.

\FloatBarrier

\begin{figure}
\begin{center}
\includegraphics[width=6in]{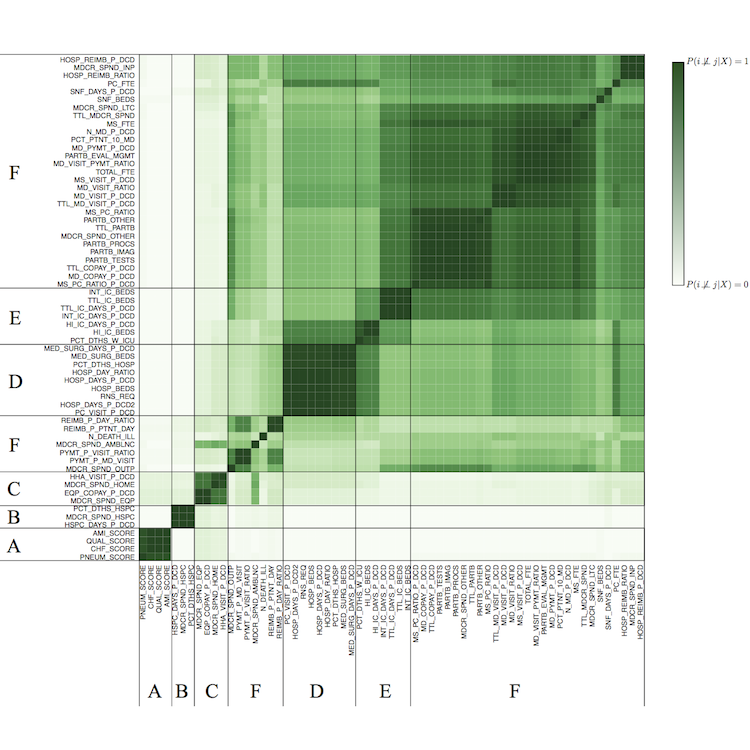}

\caption{{\bf Dependencies between variables the Dartmouth Atlas data
    aggregated by hospital referral region.} This figure shows the
  z-matrix $\mathbf{Z} = [Z_(i,j)]$ of pairwise dependence
  probabilities, where darker green represents higher
  probability. Rows and columns are sorted by hierarchical clustering
  and several portions of the matrix have been labeled. The isolation
  of [A] indicates that the quality score variables are almost
  certainly mutually dependent but independent of the variables
  describing capacity and cost structure. [B] contains three distinct
  but dependent measures of hospice cost and capacity: the percent of
  deaths in hospice, the number of hospice days per decedent, and the
  total Medicare spending on hospice usage. [C] contains spending on
  home health aides, equipment, and ambulance care. [D] shows
  dependencies between hospital stays, surgeries and in-hospital
  deaths. [E] contains variables characterizing intensive care,
  including some that probably interact with surgery, and others that
  interact with general spending metrics [F], such as usage of
  doctors' time and total full time equivalency (FTE) head count.}
\label{fig:dha_feature_z}
\end{center}
\end{figure}

The dependencies detected by CrossCat reflect accepted findings about
health care that may be surprising. The inferred pairwise dependence
probabilities, shown in Figure~\ref{fig:dha_feature_z}, depict strong
evidence for a dissociation between cost and quality. Specifically,
the variables in block A are aggregated quality scores, for congestive
heart failure ({\tt CHF\_SCORE}), pneumonia ({\tt PNEUM\_SCORE}),
acute myocardial infarction ({\tt AMI\_SCORE}), and an overall quality
metric ({\tt QUAL\_SCORE}). The probability that they depend on any
other variable in the dataset is low. This finding has been reported
consistently across multiple studies and distinct patient populations
\citep{dhasummary}. Partly due to its coverage in the popular press
\citep{gawande09}, it also informed the design of performance-based
funding provisions in the 2009 Affordable Care Act.

CrossCat identifies several other clear, coherent blocks of variables
whose dependencies are broadly consistent with common sense. For
example, Section B of Figure~\ref{fig:dha_feature_z} shows that
CrossCat has inferred probable dependencies between three variables
that all measure hospice usage. The dependencies within Section C
reflect the proposition that the presence of home health aides ---
often consisting of expensive equipment --- and overall equipment
spending are probably dependent. The dark green bar for {\tt
  MDCR\_SPND\_AMBLNC} with the variables in section C is also
intuitive: home health care easily leads to ambulance transport during
emergencies. Section D shows probable dependencies between the length
of hospital stays, hospital bed usage, and surgery. This section and
section E, which contains measures of ICU usage, are probably
predictive of the general spending metrics in section F, such as total
Medicare reimbursement, use of doctors' time, and total full time
equivalent (FTE) head count. Long hospital stays, surgery, and time in
the intensive care unit (ICU) are key drivers of costs, but not
quality of care.

\begin{figure}
\begin{center}
\includegraphics[width=6in]{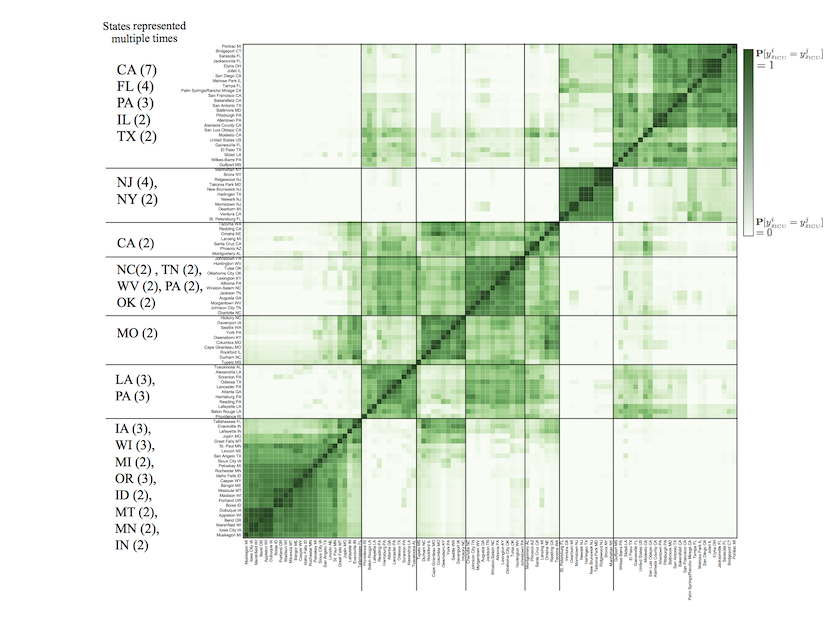}
\caption{{\bf The pairwise similarity measure inferred by CrossCat in the context of ICU utilization.} Each cell contains an estimate of the marginal probability that the hospital reporting regions corresponding to the row and column come from the same category in the view. The block structure in this matrix reflects regional variation in ICU utilization and in other variables that are probably predictive of it; examples include measures of hospital and intensive care capacity and usage. }
\label{fig:dha_entity_z}
\end{center}
\end{figure}

\begin{figure}
\begin{center}
\includegraphics[width=5in]{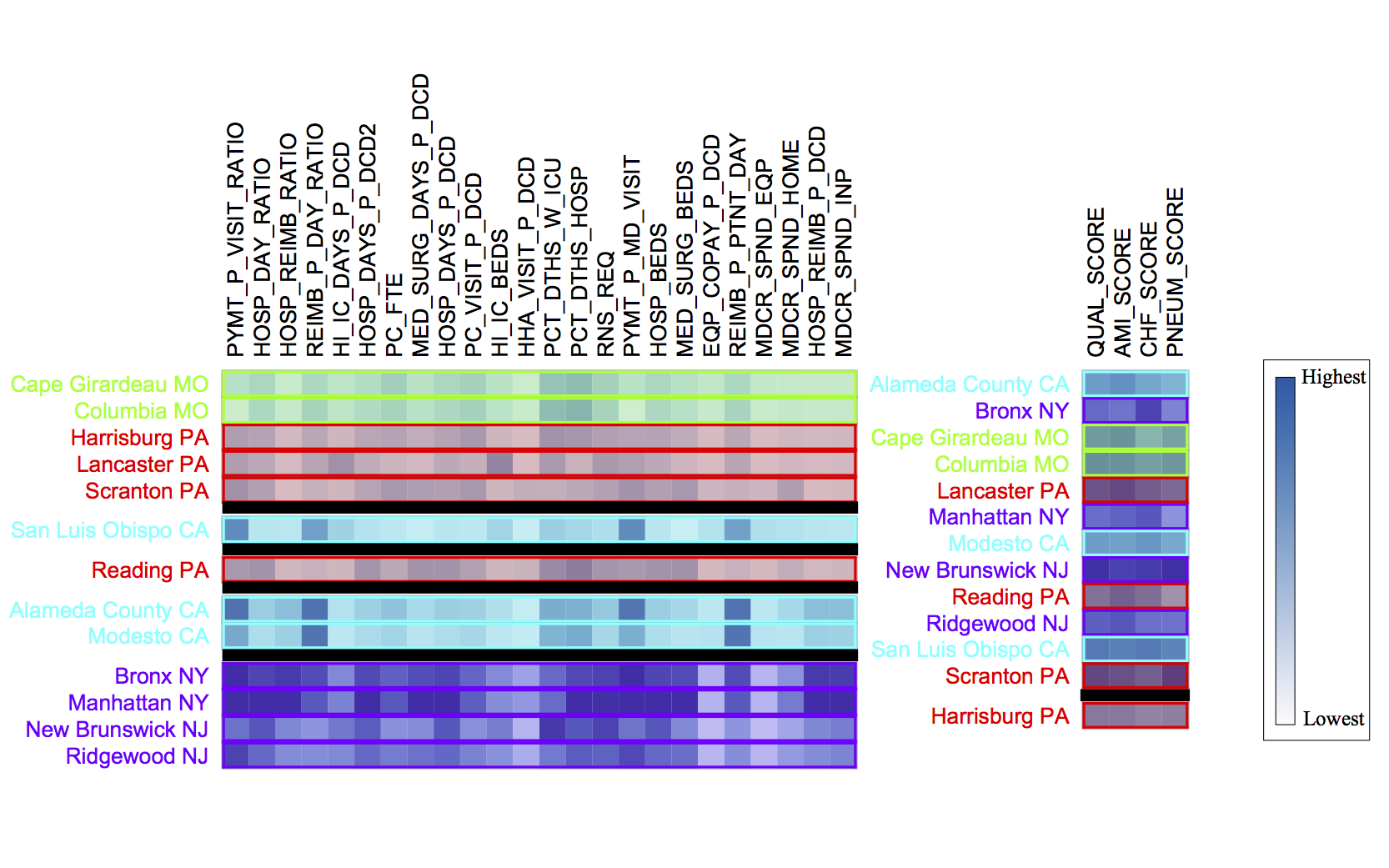} 
\caption{{\bf Subset of a single posterior sample for the Dartmouth Health Atlas.} These entities have been color-coded according to geography. Variables related to quality are independent of geography (left), but variables related to usage of services are related to geography (right). This is in accord with a key finding from \citet{gawande09}.}
\label{fig:dha_entity_z_sample}
\vspace{-4mm}
\end{center}
\end{figure}

It has been proposed that regional differences explain variation in
cost and capacity, but not quality of care \citep{gawande09}. This
proposal can be explored using CrossCat by examining individual
samples as well as the context-sensitive pairwise co-categorization
probabilities (similarities) for
hospitals. Figure~\ref{fig:dha_entity_z} shows these probabilities in
the context of time spent in the ICU. These probabilities yield
hospital groups that often contain adjacent regions, consistent with
the idea that local variation in training or technique diffusion may
contribute significantly to
costs. Figure~\ref{fig:dha_entity_z_sample} shows results for regions
from four states, coloring regions from the same state with the same
color, with white space separating categories in a given
view. Variables probably dependent on usage lead to geographically
consistent partitions, while variables that are probably dependent on
quality do not.

\begin{figure}[h]
\begin{center}
\includegraphics[width=5in]{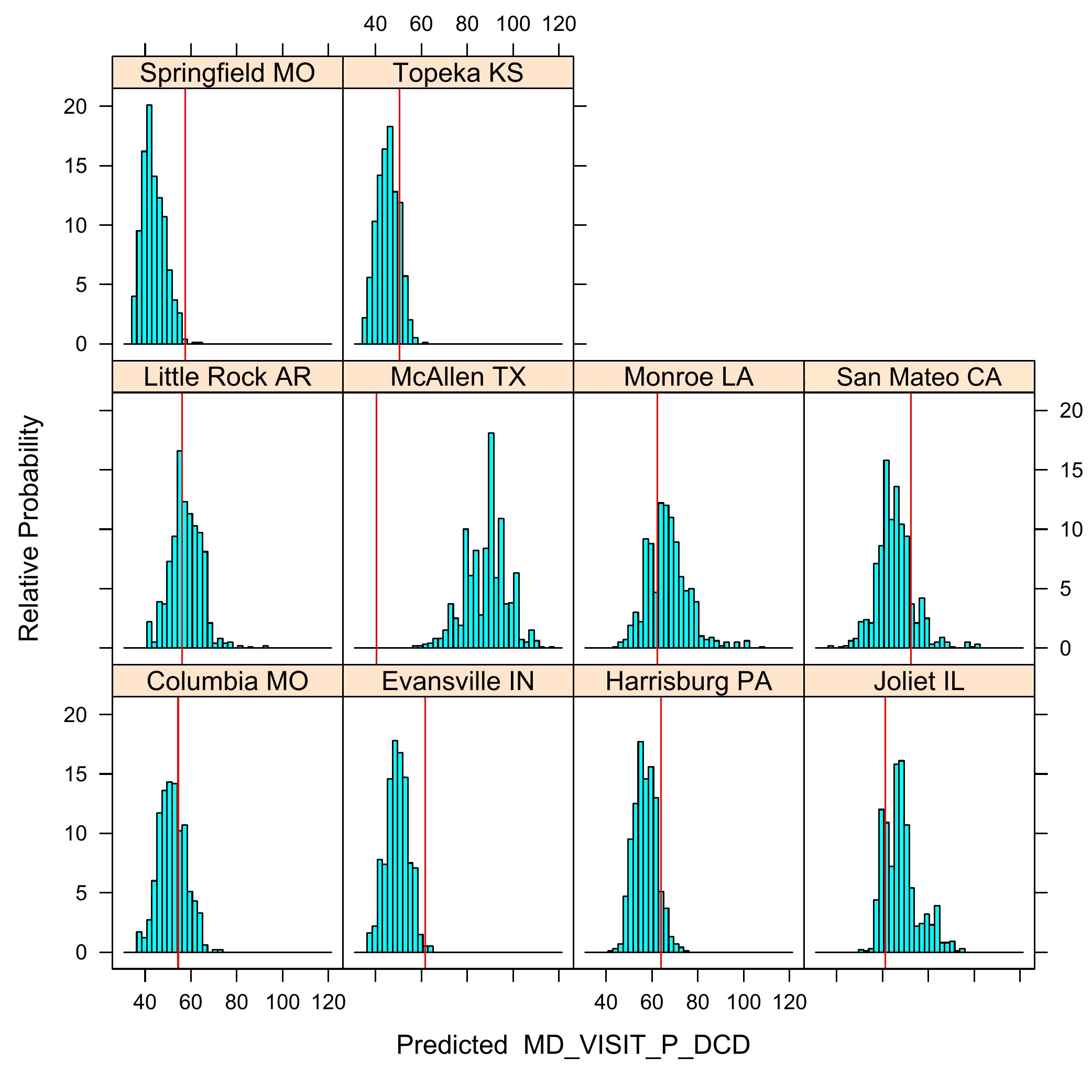}
\caption{\label{fig:dha_benchmark_icu} {\bf A comparison between the
    observed utilization of physicians their inferred predictive
    distributions, for 10 hospital reporting regions.} McAllen, TX
  appears to under-utilize physicians relative to CrossCat's
  predictions based on the overall dataset. This is consistent with
  analyses of McAllen's equipment and procedure-intensive approach to
  care. Note that some predictive distributions are multi-modal,
  e.g. Joilet, IL.}\end{center}
\end{figure}

The models inferred by CrossCat can also be used to compare each value
in the dataset with the values that are probable given the rest of the
data. Figure~\ref{fig:dha_benchmark_icu} shows the predictive
distribution on the number of physician visits for ICU patients for 10
hospital reporting regions. The true values are relatively probable
for most of these regions. However, for McAllen, Texas, the observed
value is highly improbable. McAllen is known for having exceptionally
high costs and an unusually large dependence on expensive,
equipment-based treatment rather than physician care. In fact,
\citet{gawande09} used McAllen to illustrate how significant the
variation can be.

\begin{figure}[h]
\begin{center}
\includegraphics[width=5in]{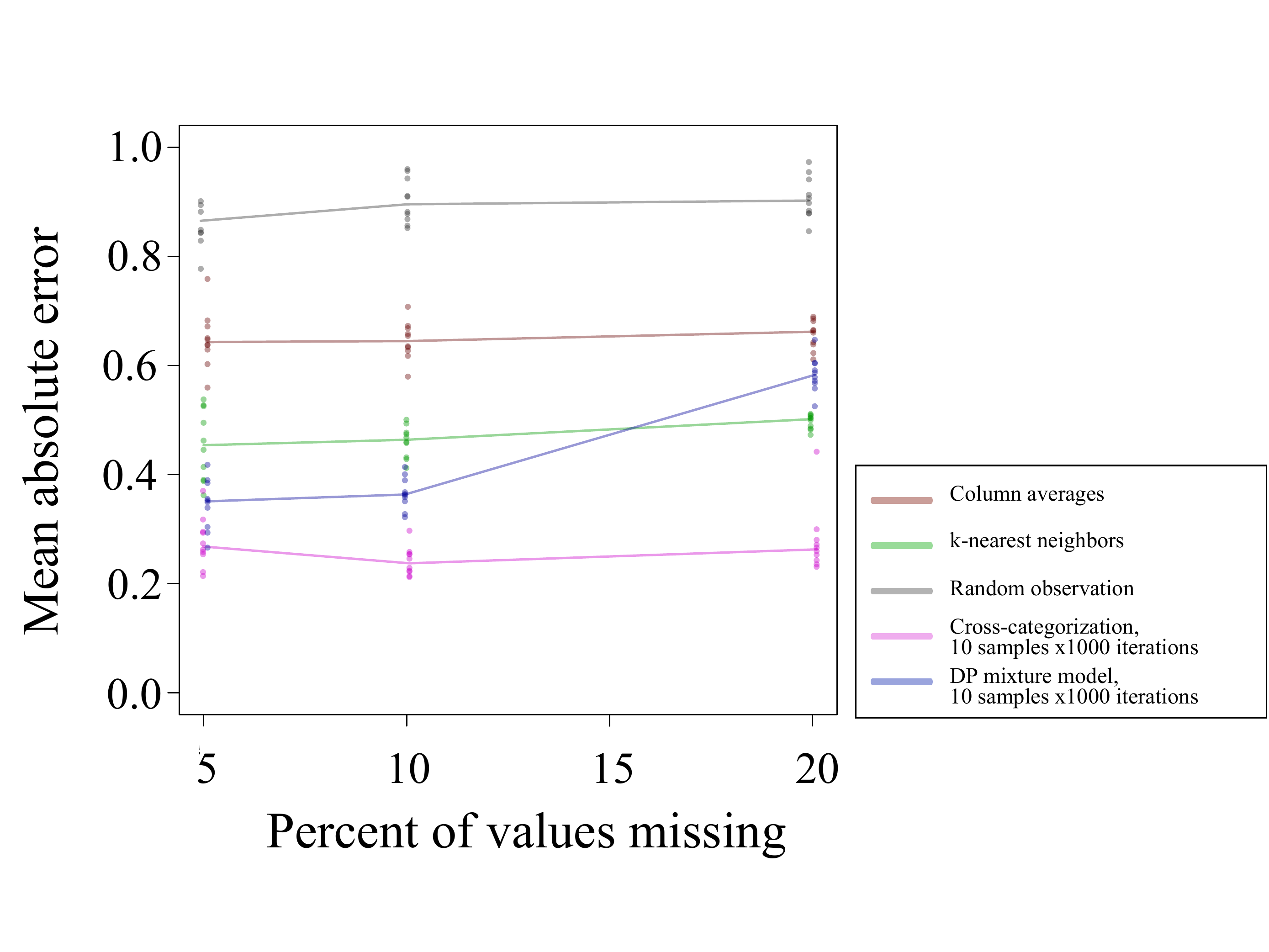}
\caption{\label{fig:dha_imputation}
{\bf A comparison of imputation accuracy under random censoring.} The error (y axis) as a function of the fraction of missing values (x axis) is measured on a scale that has been normalized by column-wise variance, so that high-variance variables do not dominate the comparison. CrossCat is more accurate than baselines such as column-wise averaging, imputation using a randomly chosen observation, a state-of-the-art variant of k-nearest-neighbors, and Dirichlet process mixtures of Gaussians. 
Also note the collapse of mixture modeling to column-wise averaging when the fraction of missing values grows sufficiently large.}
\end{center}
\end{figure}

The imputation accuracy of CrossCat on this dataset is also favorable
relative to multiple baselines. Figure~\ref{fig:dha_imputation} shows
the results on versions of the Dartmouth Atlas of Health care where
5\%-20\% of the cells are censored at random. CrossCat performs
favorably across this range, outperforming both simple model-free
schemes and nonparametric Bayesian mixture modeling.

\FloatBarrier

\subsection{Classifying Images of Handwritten Digits}

\begin{figure}[h]
\begin{center}
(a) \includegraphics[width=2in]{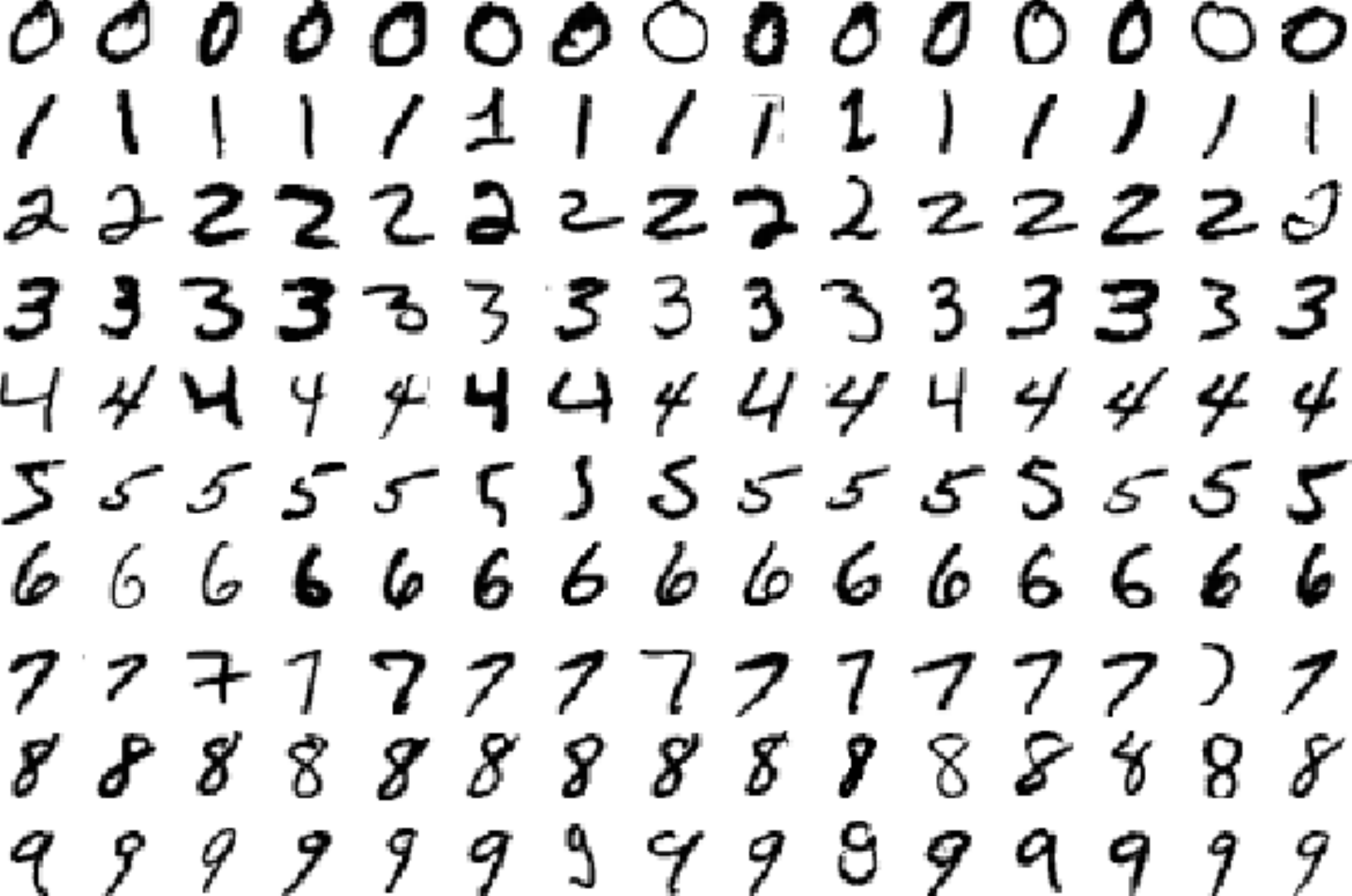}
(b) \includegraphics[width=1.4in]{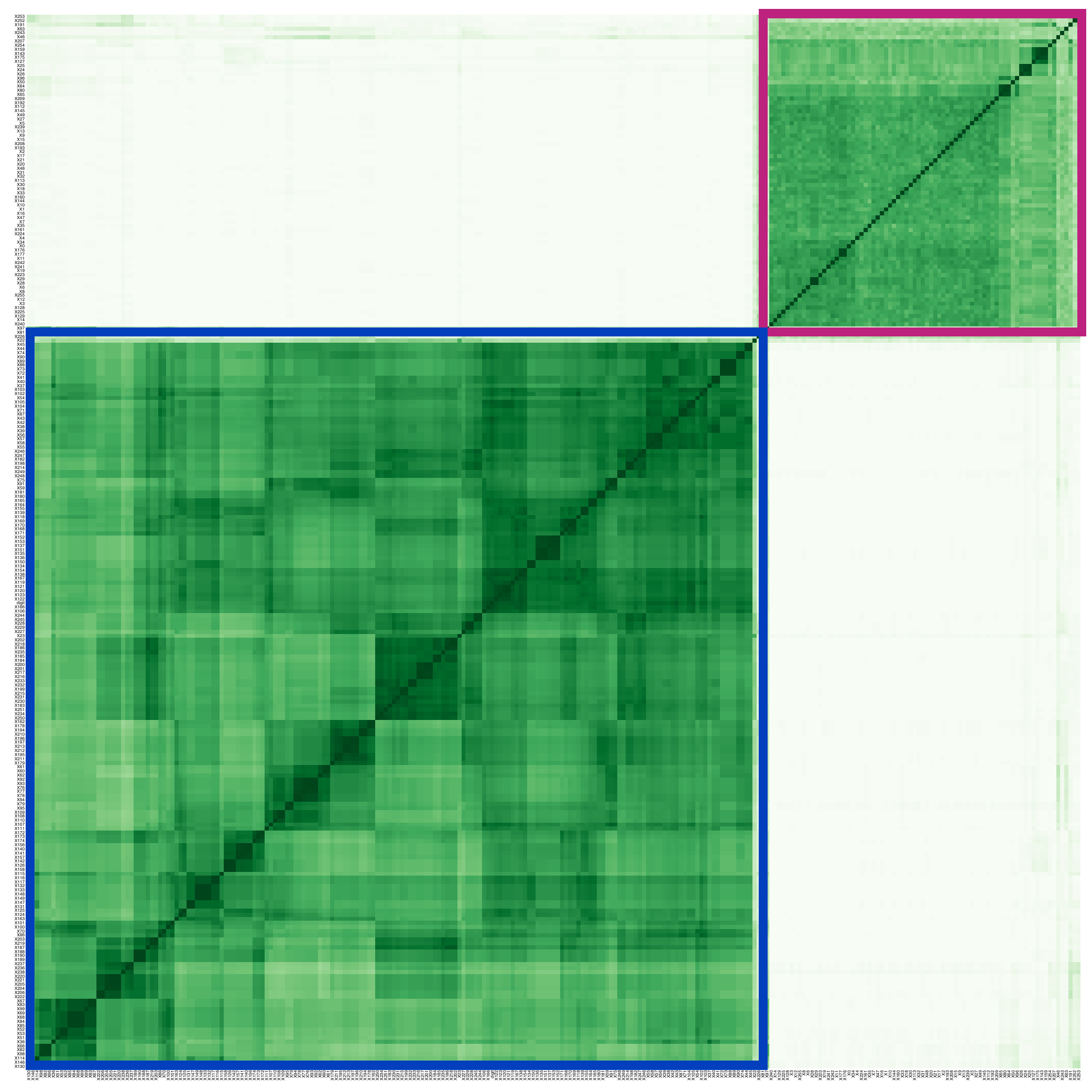}
(c) \includegraphics[width=1.5in]{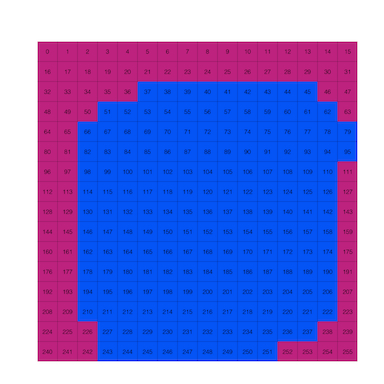}
\end{center}
\caption{
{\bf MNIST handwritten digits, feature z-matrix, and color-coded image pixel locations.}
(a) Fifteen visually rendered examples of handwritten digits for each number in the MNIST data set. Each image was converted to a binary feature vector for predictive modeling. CrossCat additionally treated the digit label as an additional feature; this value was observed for training examples and treated as missing for testing.
(b) The dependence probabilities between pixel values distinguish two blocks of pixels, one containing the digit label. (c) Coloring the pixels from each block reveals the spatial structure in pixel dependencies. Blue pixels --- pixels from the block with a blue border from figure (b) --- pick out the foreground, i.e. pixels whose values depend on what digit the image contains. Magenta pixels pick out the common background, i.e. pixels whose values are independent of what digit is drawn.}
\label{fig:mnist-z}
\end{figure}

The MNIST collection of handwritten digit images \citep{mnist} can be
used to explore CrossCat's applicability to high-dimensional
prediction problems from pattern recognition. Figure~\ref{fig:mnist-z}a
shows example digits. For all experiments, each image was downsampled
to 16x16 pixels and represented as a 256-dimensional binary
vector. The digit label was treated as an additional categorical
variable, observed for training examples and treated as missing for
testing. Figure~\ref{fig:mnist-z}b shows the inferred dependence
probabilities among pixels and between the digit label and the
pixels. The pixels that are identified as independent of the digit
class label lie on the boundary of the image, as shown in
Figure~\ref{fig:mnist-z}c.

\begin{figure}
\begin{center}
\includegraphics[width=6in]{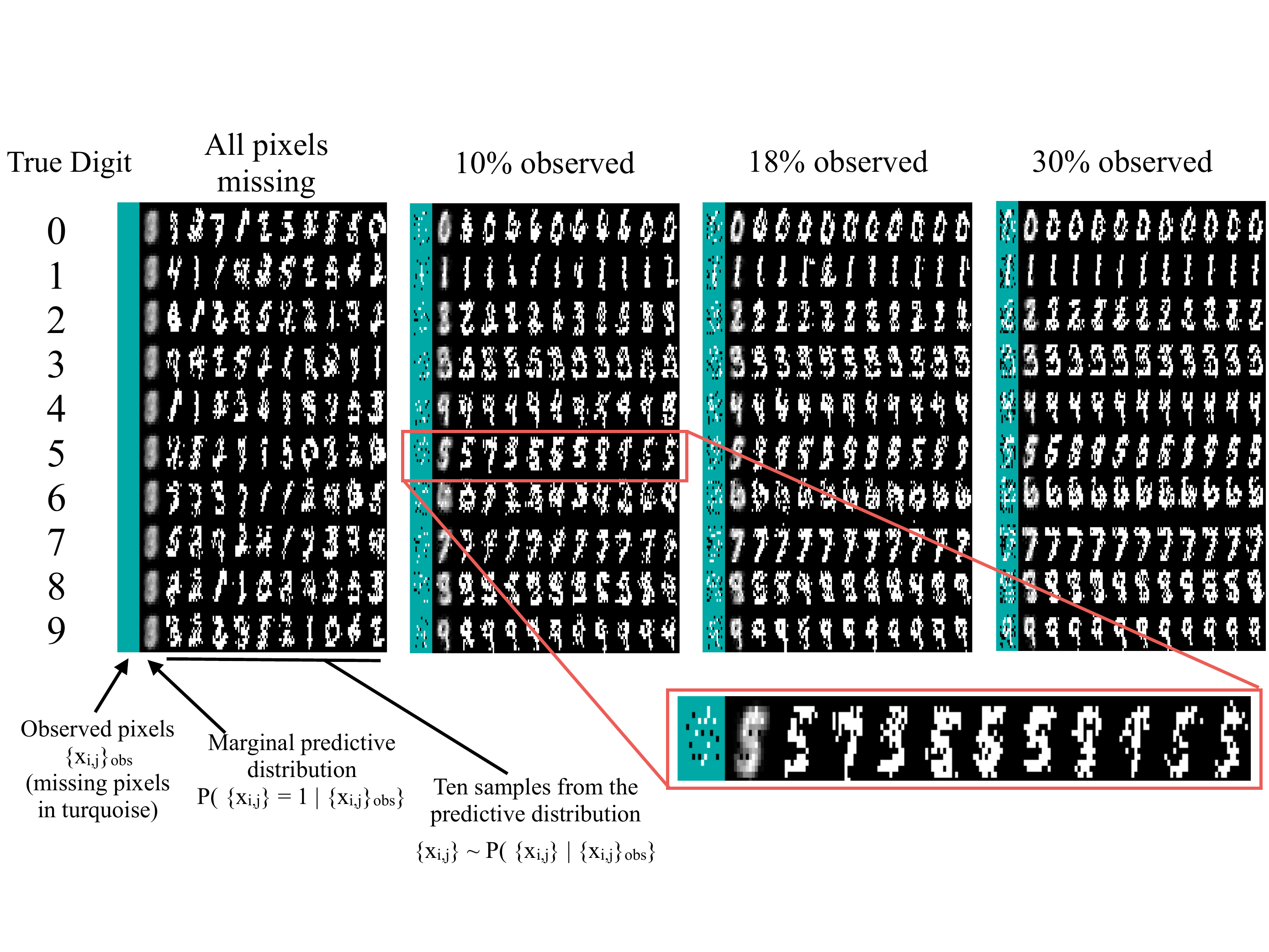}
\end{center}
\caption{{\bf Predicted images of handwritten digits given sparse
    observations.} CrossCat can be used to fill in missing pixels by
  sampling from their conditional density given the observed
  pixels. Each panel shows completion results for one image per digit;
  across panels, the fraction of observed pixels grows from 0 to
  30\%. The leftmost column shows the observed pixels in black and
  white, with missing pixels in turquoise. The second column from the
  left shows the marginal probabilities for each pixel. The 10
  remaining columns show independent sampled predictions. In the
  leftmost panel, with no data, all 10 predictive distributions are
  equivalent. The predictive distribution collapses onto single digit
  classes (except for 8 and 6) after 18\% of the digits have been
  observed. The marginal images for 1, 7, and 9 become resolvable to a
  single prototypical example after 10\% of the 256 pixels have been
  observed. Others, such as 8, remain ambiguous.}
\label{fig:mnist-gen}
\end{figure}

A set of approximate posterior samples from CrossCat can be used to
complete partially observed images by sampling predictions for
arbitrary subsets of pixels. Figure~\ref{fig:mnist-gen} illustrates
this: each panel shows the data, marginal predictive images, and
predicted image completions, for 10 images from the dataset, one per
digit. With no data, all 10 predictive distributions are equivalent,
but as additional pixels are observed, the predictions for most images
concentrate on representations of the correct digit. Some digits
remain ambiguous when $\sim$30\% of the pixels have been observed. The
predictive distributions begin as highly multi-modal distributions
when there is little to no data, but concentrate on roughly unimodal
distributions given sufficiently many features.

\begin{figure}
\begin{tabular}{ccc}
\includegraphics[width=2in]{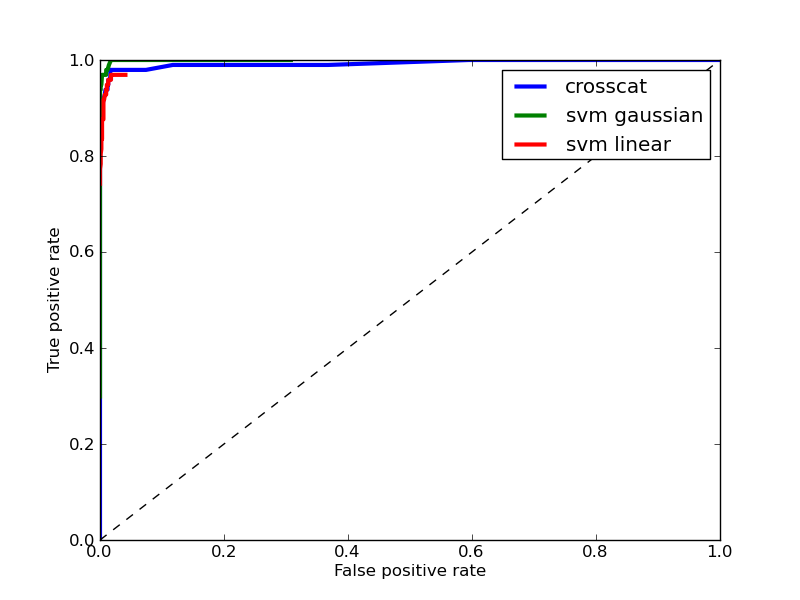} &
\includegraphics[width=2in]{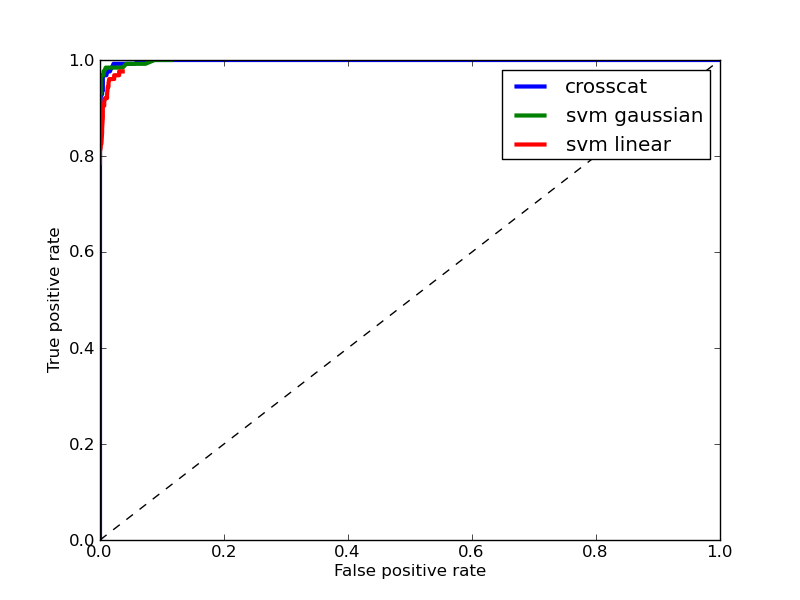} &
\includegraphics[width=2in]{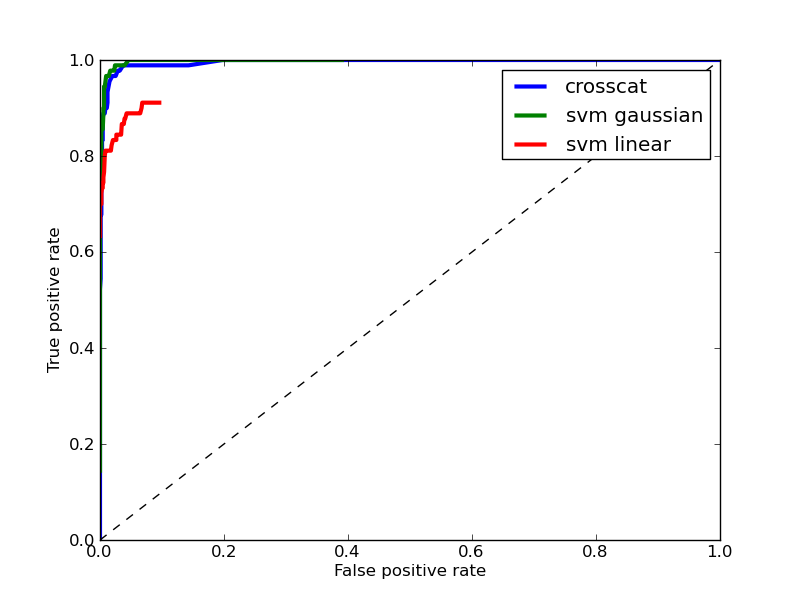} \\
\includegraphics[width=2in]{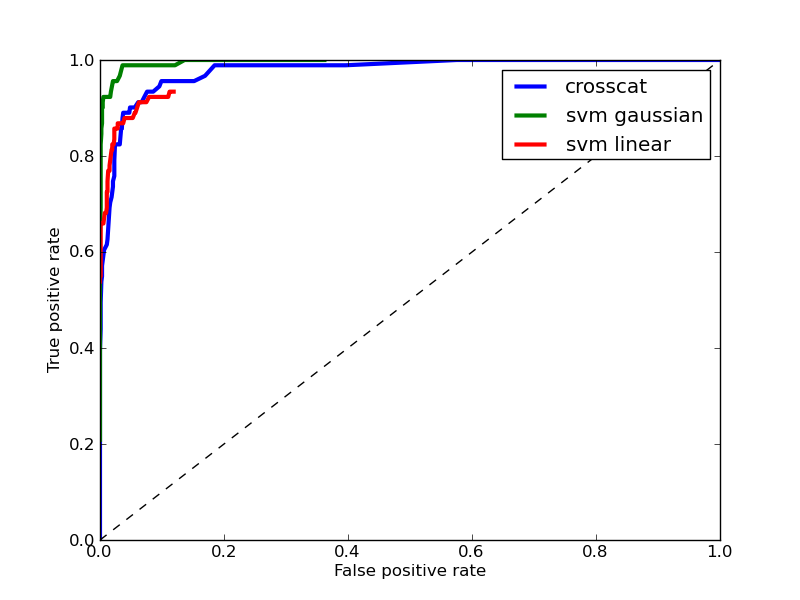} &
\includegraphics[width=2in]{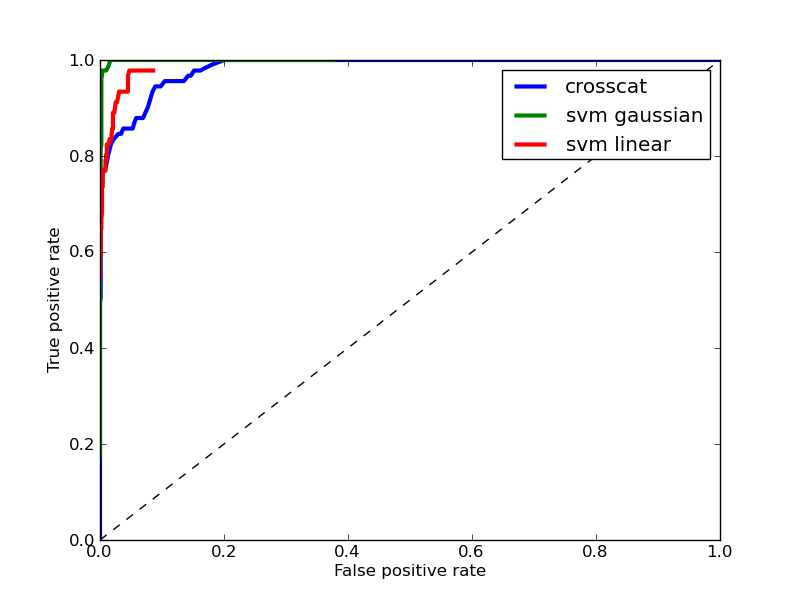} &
\includegraphics[width=2in]{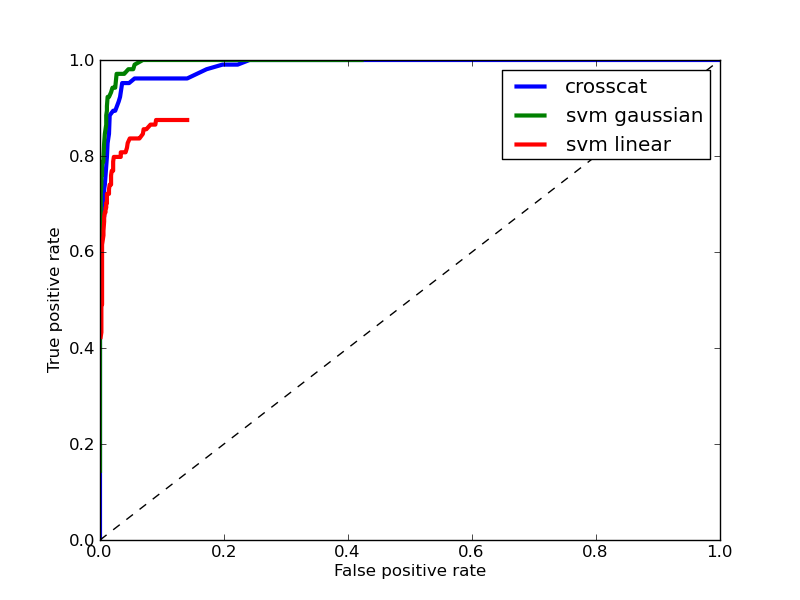} \\
\includegraphics[width=2in]{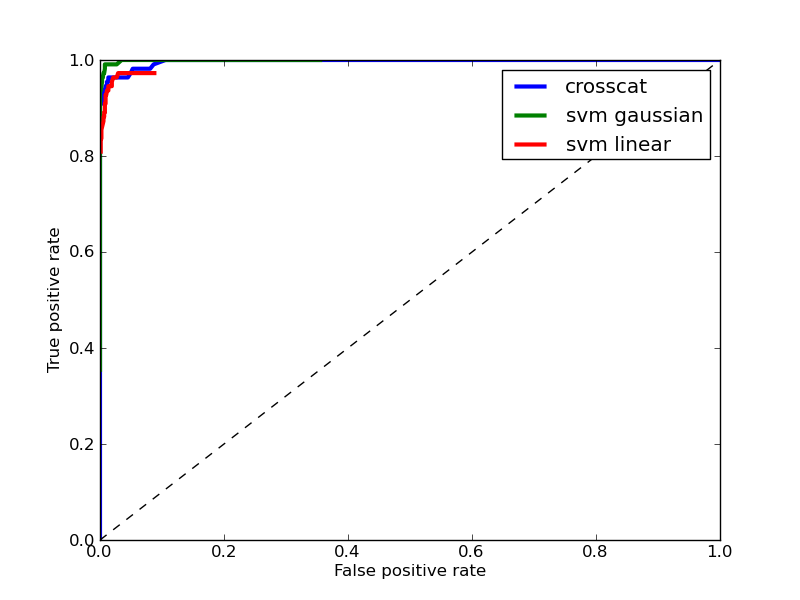} &
\includegraphics[width=2in]{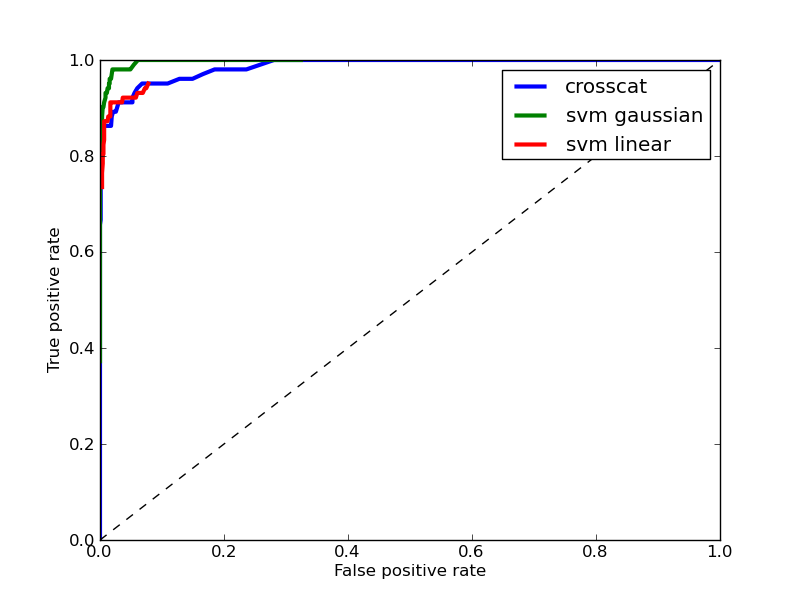} &
\includegraphics[width=2in]{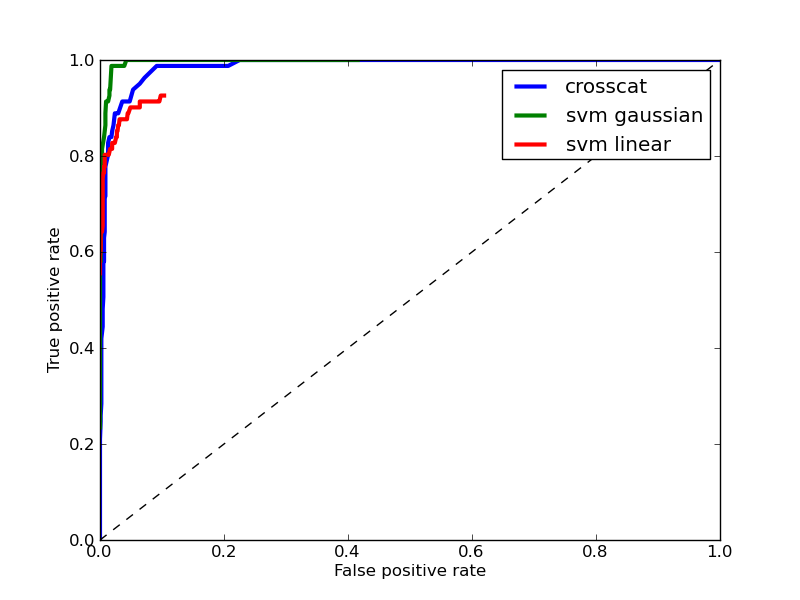} \\
\includegraphics[width=2in]{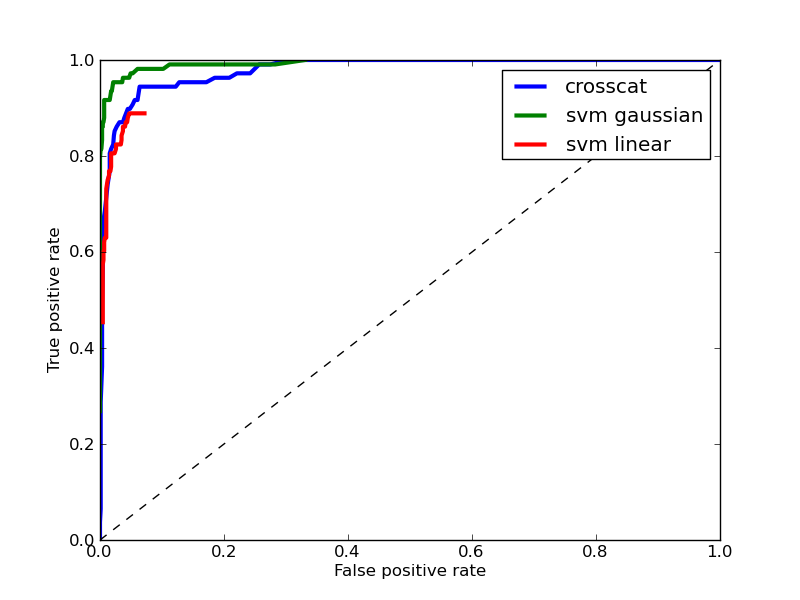} \\
\end{tabular}
\caption{{\bf Classification accuracy on handwritten digits from
    MNIST.} Each panel shows the true positive/false positive tradeoff
  curves for classifying each digit from 0 through 9. Digit images
  were represented as binary vectors, with one dimension per pixel. As
  with the image completion example from Figure~\ref{fig:mnist-gen},
  CrossCat was applied directly to this data, with the digit label
  appended as a categorical variable; no weighting or tuning for the
  supervised setting was done. Support vector machines (SVMs) with
  both linear and Gaussian kernels are provided as
  baselines. Regularization and kernel bandwidth parameters were
  chosen via cross-validation on 10\% of the training data, with
  multiple classes treaded via a one-versus-all reduction. See the
  main text for further discussion.}
\label{fig:mnist-roc}
\end{figure}

The predictive distribution can also be used to infer the most
probable digit, i.e. solve the standard MNIST multi-class
classification problem. Figure~\ref{fig:mnist-roc} shows ROC curves
for CrossCat on this problem. Each panel shows the tradeoff between
true and false positives for each digit, aggregated from the overall
performance on the underlying multi-class problem. The figure also
includes ROC curves for Support Vector Machines with linear and
Gaussian kernels. For these methods, the standard one-vs-all approach
was used to reduce the multi-class problem into a set of binary
classification problems. The regularization and kernel bandwidth
parameters for the SVMs were set via cross-validation using 10\% of
the training data. 10 posterior samples from CrossCat were used, each
obtained after 1,000 iterations of inference from a random
initialization. CrossCat was more accurate than the linear
discriminative technique; this is expected, as CrossCat induces a
nonlinear decision boundary even if classifying based on a single
posterior sample. Overall, the 10-sample model used here made less
accurate predictions than the Gaussian SVM baseline. Also, in
anecdotal runs that were scored by overall 0-1 loss rather than
per-digit accuracy, performance was similarly mixed, and less
favorable for CrossCat. However, the size of the kernel matrix for the
Gaussian SVM scales quadratically, while CrossCat scales linearly. As
a classifier, CrossCat thus offers different tradeoffs between
accuracy, amount of training data, test-time parallelism (via the
number of independent samples), and latency than standard techniques.

\FloatBarrier

\subsection{Voting records for the 111th Senate}

Voters are often members of multiple issue-dependent coalitions. For
example, US senators sometimes vote according to party lines, and at
other times vote according to regional interests. Because this
common-sense structure is typical for the CrossCat prior, voting
records are an important test case.

\begin{figure}[t]
\centering
\includegraphics[width=7in]{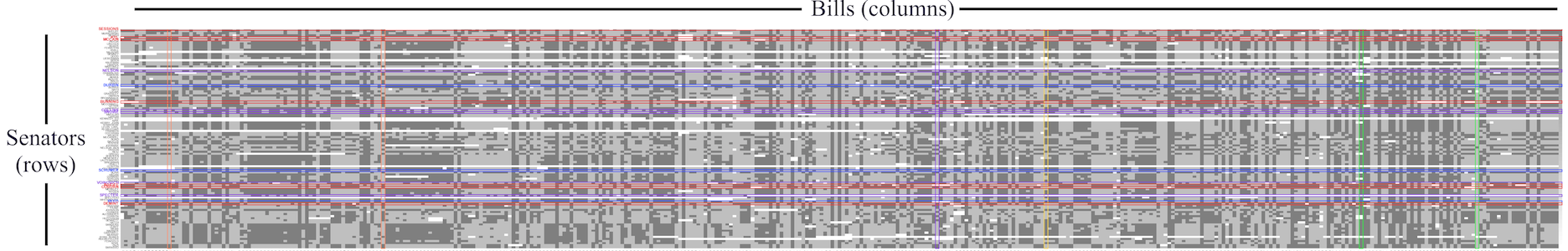}
\includegraphics[width=7in]{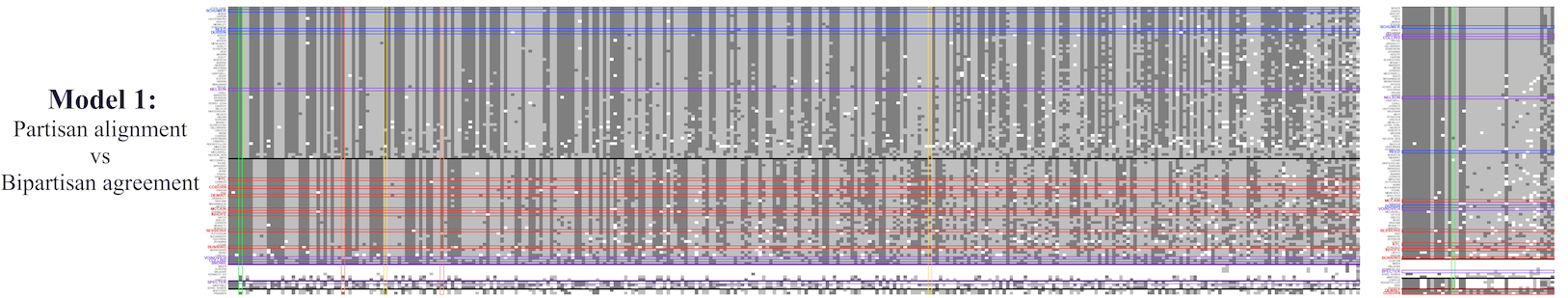}
\includegraphics[width=7in]{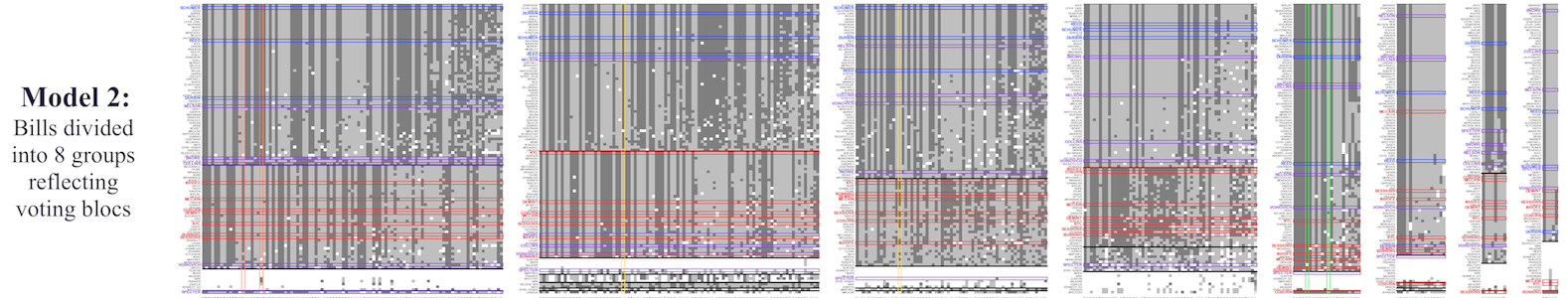}
\caption{{\bf Voting records for the 111th US Senate (2009).} (top) This includes 397 votes (yea in light grey, nay in dark grey) for 106 senators, including separate records for senators who changed parties or assumed other offices mid-term. Some senators are highlighted in colors based on their generally accepted identification as democrats (blue), moderates (purple), or republicans (red). See main text for an explanation of the colored bills.
(middle) This row shows a simple or “low resolution” posterior sample that divides bills into those that exhibit partisan alignment and those with bipartisan agreement. Clusters of senators, generated automatically, are separated by thick black horizontal lines.
(bottom) This row shows a sample that includes additional views and clusters, positing a finer-grained predictive model for votes that are treated as random in the middle row.}
\label{fig:senate_data}
\end{figure}

This set of experiments describes the results of a CrossCat analysis
of the 397 votes held by the 111th Senate during the 2009-2010
session. In this dataset, each column is a vote or bill, and each row
is a senator. Figure~\ref{fig:senate_data} shows the raw voting data,
with several votes and senators highlighted. There are 106 senators;
this is senators is larger than the size of the senate by 6, due to
deaths, replacement appointments, party switches, and special
elections. When a senator did not vote on a given issue, that datum is
treated as missing. Figure~\ref{fig:senate_data} also includes two
posterior samples, one that reflects partisan alignment and another
that posits a higher-resolution model for the votes.

\begin{figure}[t]
\includegraphics[width=6in]{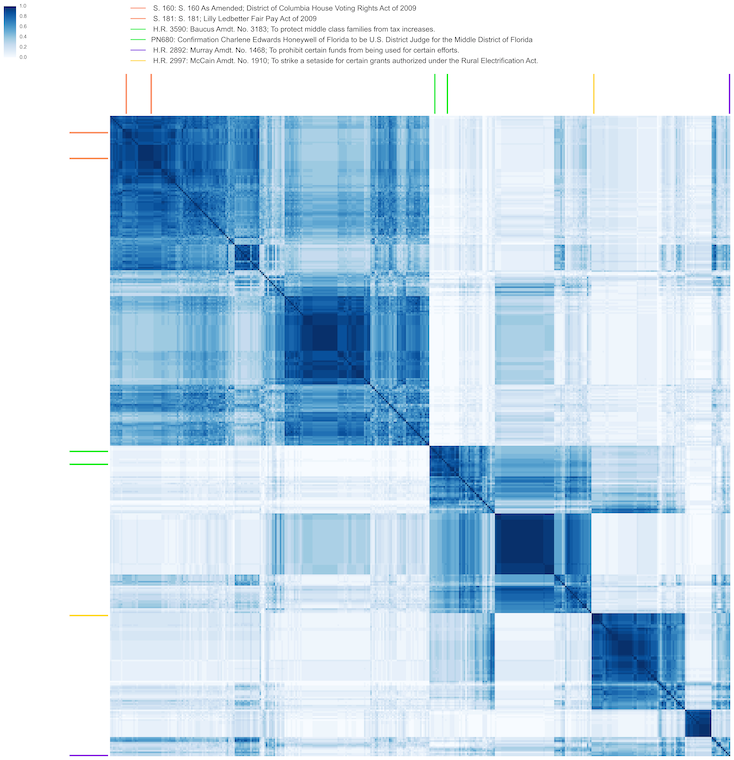}
\caption{{\bf Pairwise dependence probabilities between bills.} Blocks
  of bills with high probability of dependence include predominantly
  partisan issues (orange), issues with broad bipartisan support
  (green), and bills that divide senators along ideological or
  regional lines.}
\label{fig:senate_z}
\end{figure}

\begin{figure}[t]
\includegraphics[width=3in]{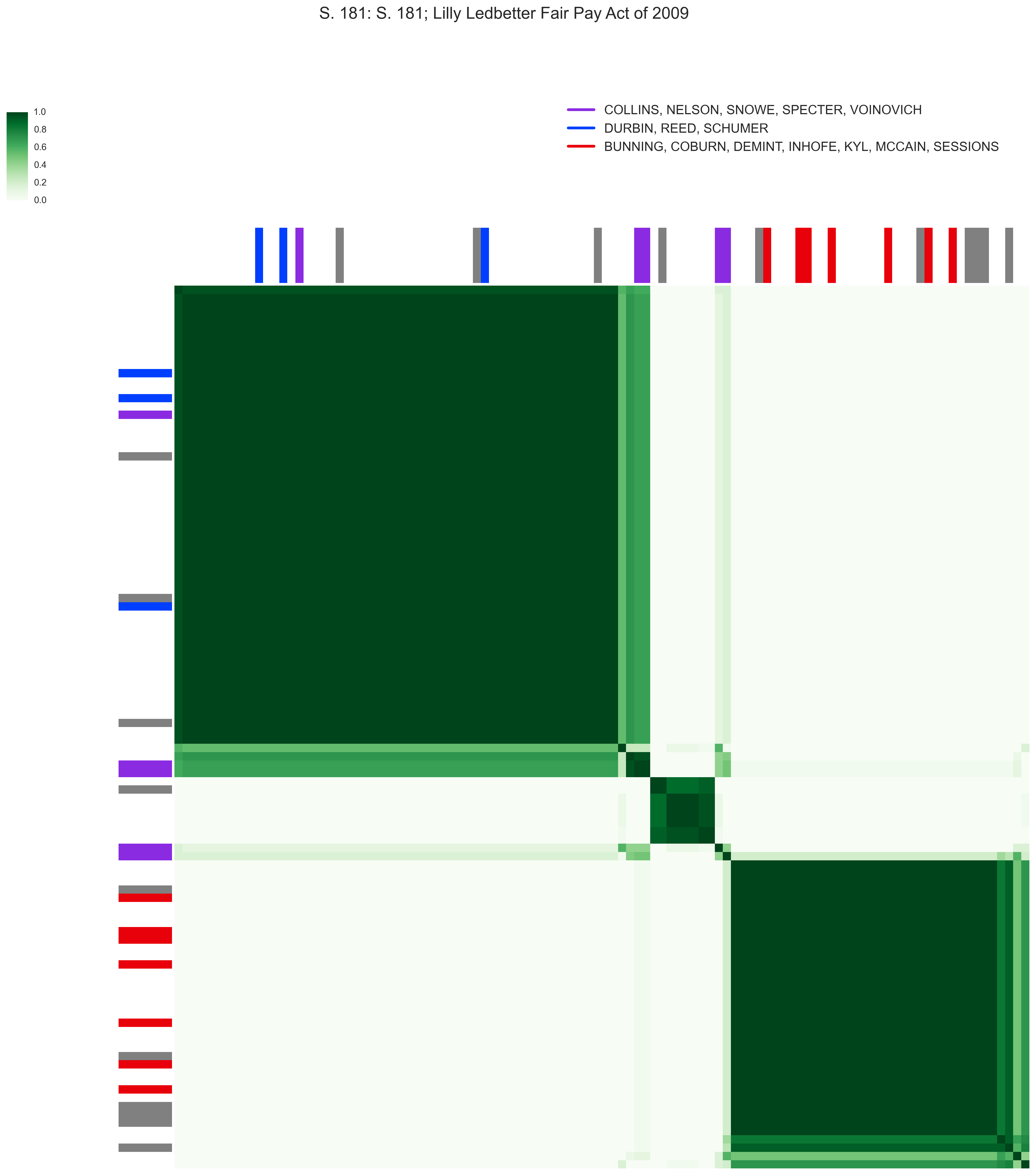}
\includegraphics[width=3in]{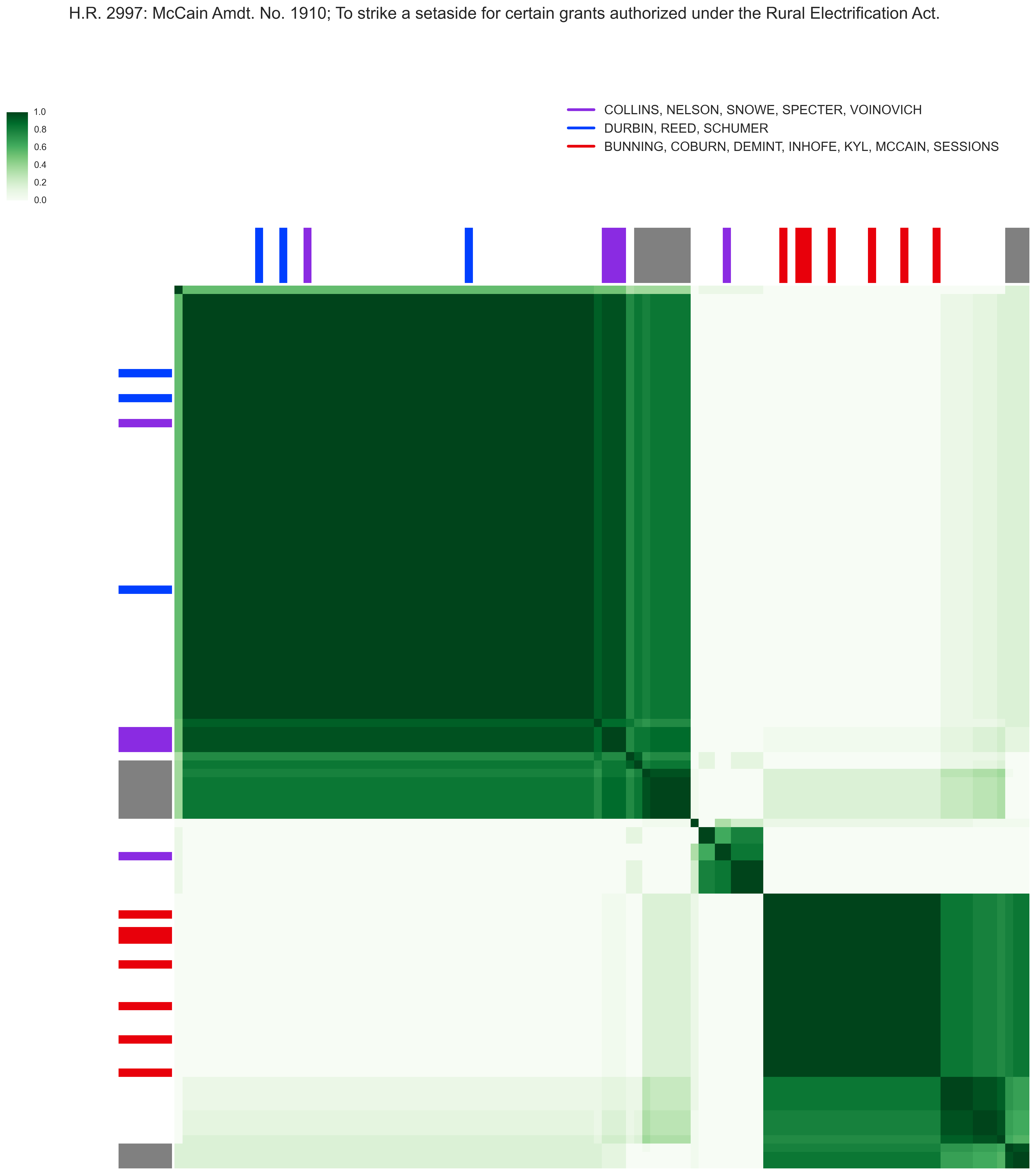}
\caption{{\bf Context-sensitive similarity measures for senators with respect to partisan and special-interest issues.} The left matrix shows senator similarity for S. 181, the Lilly Ledbetter Fair Pay Act of 2009, a bill whose senator clusters tend to respect party lines. The right matrix is for H.R. 2997, a bill designed to remove a subsidy for energy generation systems in rural areas. The grey senators are those whose similarities changed the most between these two bills.}
\label{fig:senate_similarity}
\end{figure}

This kind of structure is also apparent in estimates that aggregate
across samples. Dependence probabilities between votes are shown in
Figure~\ref{fig:senate_z}. The visible independencies between blocks
are compatible with a common-sense understanding of US politics. The
two votes in orange are partisan issues. The two votes in green have
broad bipartisan support. The vote in yellow aimed at removing an
energy subsidy for rural areas, an issue that cross-cuts party
lines. The vote in purple stipulates that the Department of Homeland
Security must spend its funding through competitive processes, with an
exception for small businesses and women or minority-owned
businesses. This issue subdivides the republican party, isolating many
of the most fiscally conservative. Similarity matrices for the
senators with respect to S. 160 (orange) and an amendment to H.R. 2997
(yellow) are shown in Figure~\ref{fig:senate_similarity}, with the
senators whose similarity values changed the most between these two
bills highlighted in grey.

\begin{figure}[t]
\begin{center}
(a) \includegraphics[width=1.8in]{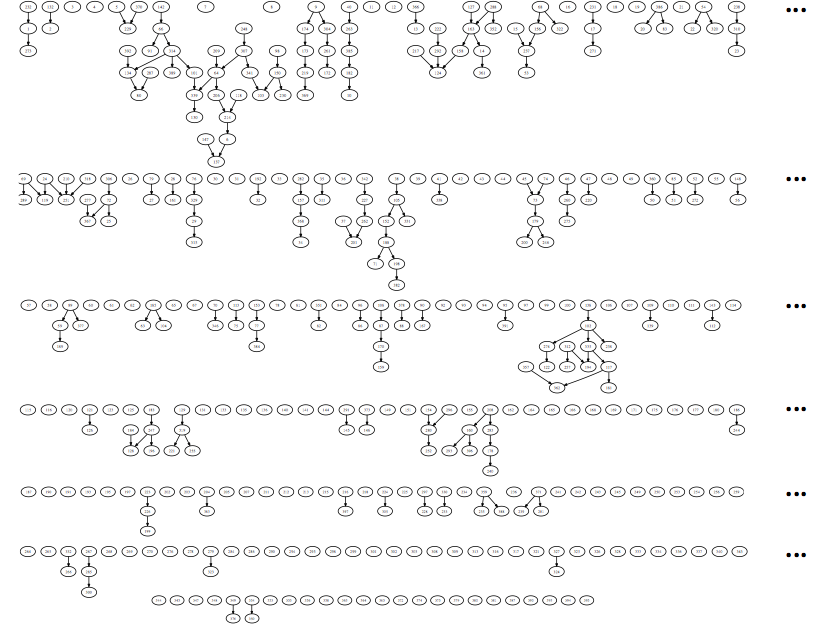}
(b) \includegraphics[width=1.5in]{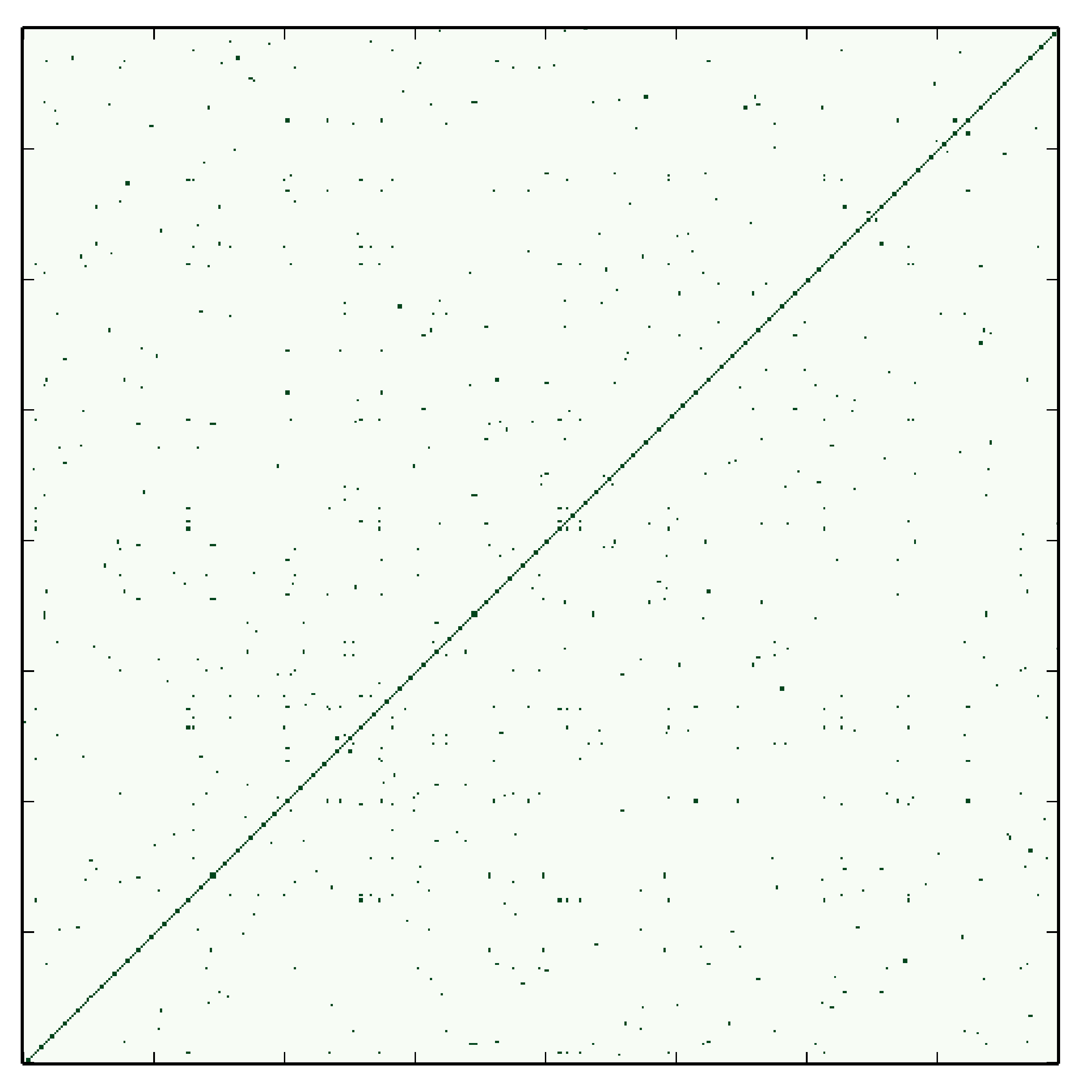}
(c) \includegraphics[width=1.8in]{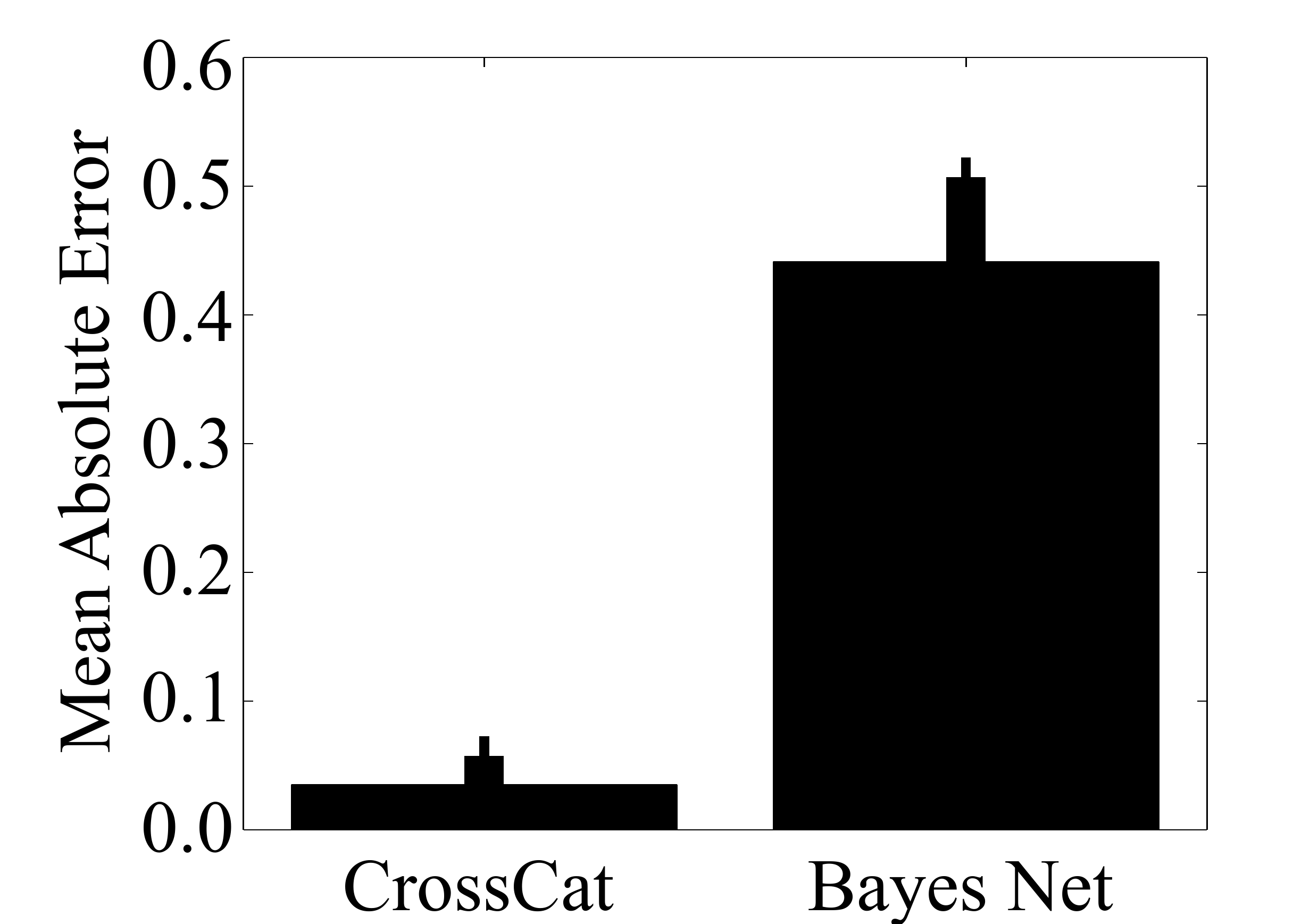}
\caption{ {\bf Comparison of latent structures and predictive accuracy
    for CrossCat and Bayesian network structure learning.} (a) The
  Bayesian network structure found by structure learning; each node is
  a vote, and edge indicates a conditional dependence. (b) The sparse
  marginal dependencies induced by this Bayes net. (c) A comparison of
  the predictive accuracy of CrossCat and Bayesian networks. See main
  text for details and discussion.}
\label{fig:senate-cc-bnet}
\end{center}
\end{figure}

It is instructive to compare the latent structure and predictions
inferred by CrossCat with structures and predictions from other
learning techniques. As an individual voting record can be described
by 397 binary variables and the missing values are negligible,
Bayesian network structure learning is a suitable
benchmark. Figure~\ref{fig:senate-cc-bnet}a shows the best Bayesian
network structure found by structure learning using the
search-and-score method implemented in the Bayes Net Toolbox for
MATLAB \citep{Murphy01thebayes}. This search is based on local moves
similar to the transition operators from \citet{giudiciGreen99}. The
highest scoring graphs after 500 iterations contained between 143 and
193 links. Figure~\ref{fig:senate-cc-bnet}b shows the marginal
dependencies between votes induced by this Bayesian network; these are
sparser than those from CrossCat. Figure~\ref{fig:senate-cc-bnet}c
shows the mean absolute errors for this Bayes net and for CrossCat on
a predictive test where 25\% of the votes were held out and predicted
for senators in the test set. Bayes net CPTs were estimated using a
symmetric Beta-Bernoulli model with hyper-parameter
$\alpha=1$. CrossCat predictions were based on four samples, each
obtained after 250 iterations of inference. Compared to Bayesian
network structure learning, CrossCat makes more accurate predictions
and also finds more intuitive latent structures.

\FloatBarrier

\begin{figure}[h]
\hspace{-0.6in}
\begin{minipage}{4.5in}%
\includegraphics[width=4.5in]{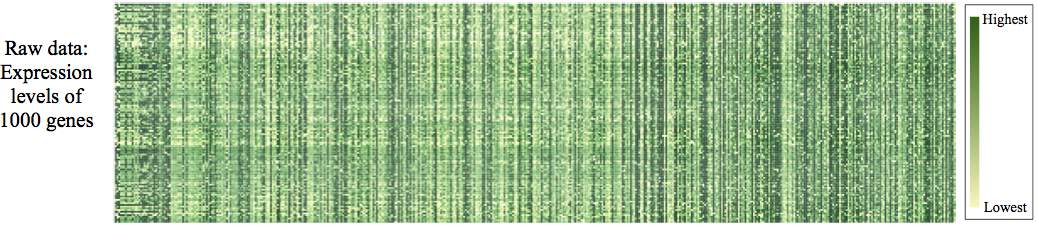} \includegraphics[width=4.5in]{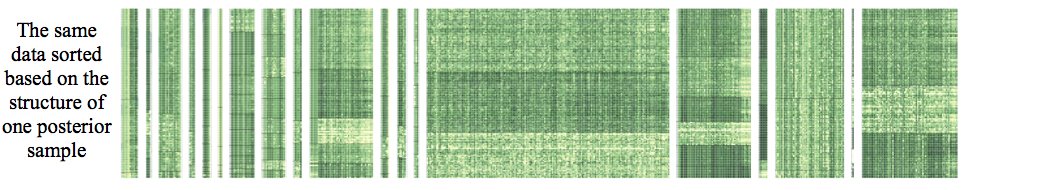}
\end{minipage}%
\begin{minipage}{3in}%
\includegraphics[width=3in]{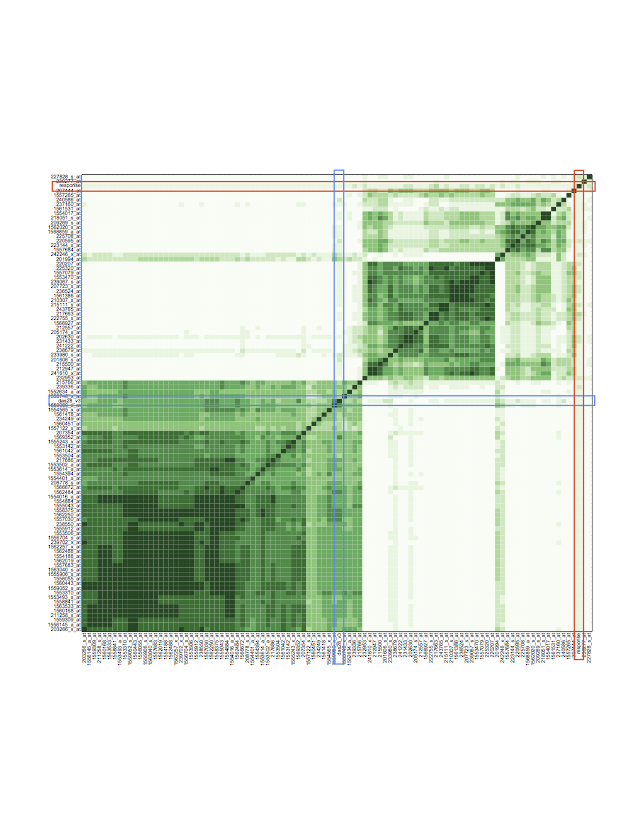}
\end{minipage}
\vspace{-0.5in}
\caption{{\bf CrossCat inference results on gene expression data.}
  (left) Expression levels for the top 1,000 highest-variance genes
  from \citet{raponi2009microrna} in original form and in the order
  induced by a single CrossCat sample. CrossCat was applied to a
  subset with roughly 1 million cells ($\sim$10,000 probes by
  $\sim$100 tissue samples). (right) Pairwise dependence probabilities
  between probe values and treatment response inferred from the
  GSE15258 dataset. The 100 probes most probably dependent on a
  3-class treatment response variable, based on analysis of a subset
  with 1,000 probes, are shown outlined in red. The low inferred
  dependence probability suggests that the data does not support the
  existence of any prognostic biomarker. A standard disease activity
  score is shown outlined in blue; this measure is naturally dependent
  on many of the probes.}
\label{fig:genes}
\end{figure}

\subsection{High-dimensional Gene Expression Data}

High-throughput measurement techniques in modern biology generate
datasets of unusually high dimension. The number of variables
typically is far greater than the sample size. For example,
microarrays can be used to assess the expression levels of 10,000 to
100,000 probe sequences, but due to the cost of clinical data
acquisition, typical datasets may have just tens or hundreds of tissue
samples. Exploration and analysis of this data can be both
statistically and computationally challenging.

Individual samples from CrossCat can aid exploration of this kind of
data. Figure~\ref{fig:genes} shows one typical sample obtained from
the data in \citet{raponi2009microrna}. The dataset had $\sim$1
million cells, with 10,000 probes (columns) and 100 tissue samples
(rows). This sample has multiple visually coherent views with $\sim$50
genes, each reflecting a particular low-dimensional pattern. Some of
these views divide the rows into clusters with ``low'', ``medium'' or
``high'' expression levels; others reflect more complex patterns. The
co-assignment of many probes to a single view could indicate the
existence of a latent co-regulating mechanism. The structure in a
single sample could thus potentially inform pathway searches and help
generate testable hypotheses.

Posterior estimates from CrossCat can also facilitate
exploration. CrossCat was applied to estimate dependence probabilities
for a subset of the arthritis dataset from
\citep{bienkowska2009convergent} (NCBI GEO accession number
GSE15258). This dataset contains expression levels for $\sim$55,000
probes, each measured for 87 patients. It also contains standard
measures of initial and final disease levels and a categorical
``response'' variable with 3 classes. CrossCat was applied to subsets
with 1,000 and 5,000 columns.

Figure~\ref{fig:genes} shows the 100 variables most probably
predictive of a 3-class treatment response variable. The dependence
probabilities with response (outlined in red) are all low,
i.e. according to CrossCat, there is little evidence in favor of the
existence of any prognostic biomarker. At first glance this may seem
to contradict \citep{bienkowska2009convergent}, which reports 8-gene
and 24-gene biomarkers with prognostic accuracies of
83\%-91\%. However, the test set from \citep{bienkowska2009convergent}
has 11 out-of-sample patients, 9 of whom are responders. Predicting
according to class marginal probabilities would yield compatible
accuracy. The final disease activation level, outlined in blue, does
appear within the selected set of variables. CrossCat infers that it
probably depends on many other gene expression levels; these genes
could potentially reflect the progression of arthritis in
physiologically or clinically useful ways.

\FloatBarrier

\subsection{Longitudinal Unemployment Data by State}

\begin{figure}[t]
\begin{center}
\includegraphics[width=7in]{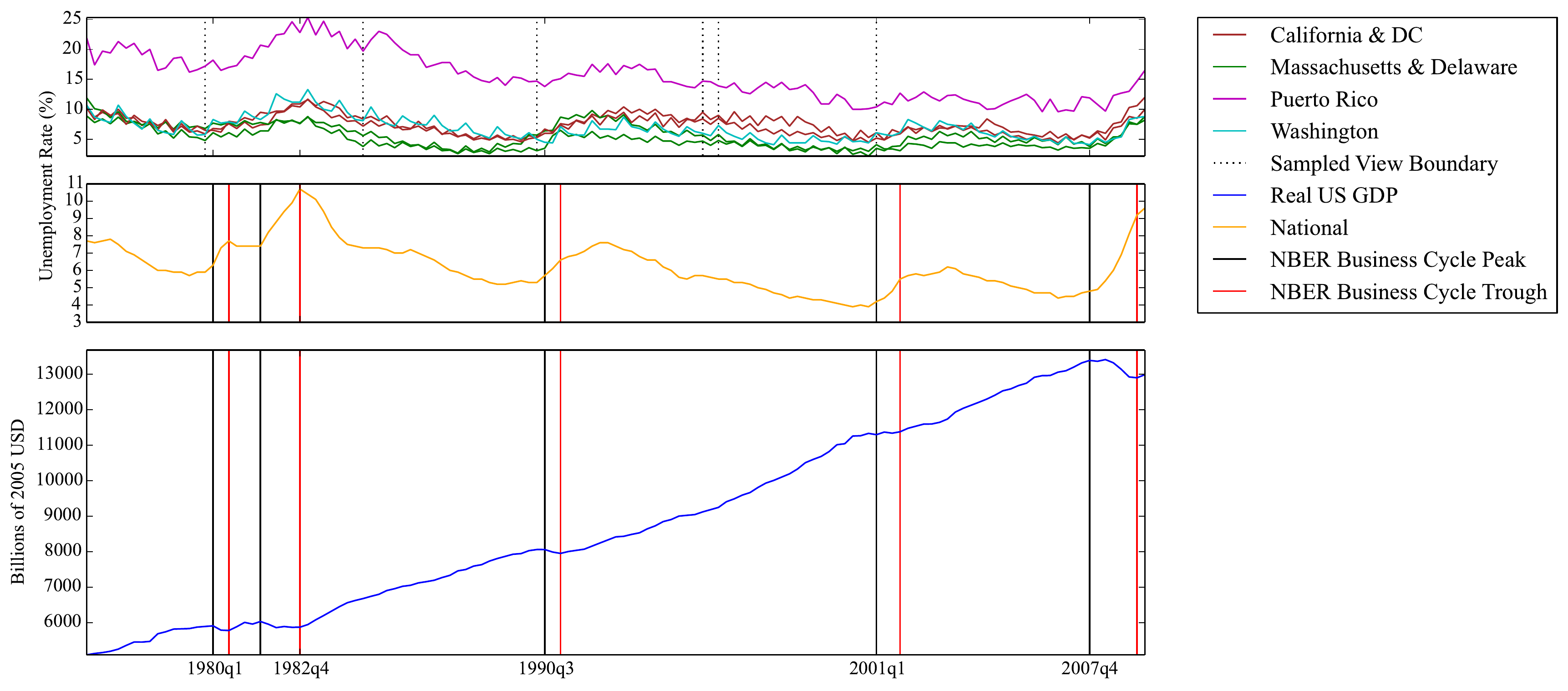}
\caption{{\bf US state-level unemployment data aligned with official business cycle peaks and troughs.} (top) Unemployment rates for five states from 1976 to 2009, colored according to one posterior sample. (middle) The national unemployment rate and business cycle peaks and troughs during the same period. Business cycle peaks and troughs are identified using multiple macroeconomic signals; see main text for details. Unemployment grows during recessions (intervals bounded on the left by black and on the right by red) and shrinks during periods of growth (intervals bounded on the left by red and on the right by black). (bottom) Real US gross domestic product (GDP) similarly decreases or stays constant during recessions and increases during periods of growth. View boundaries from the CrossCat sample pick out the business cycle turning points around 1980, 1990 and 2001.}
\label{fig:unemployment_clusters}
\end{center}
\end{figure}

This experiment explores CrossCat's behavior on data where variables
are tracked over time. The data are monthly state-level unemployment
rates from 1976 to 2011, without seasonal adjustment, obtained from
the US Bureau of Labor Statistics. The data also includes annual
unemployment rates for every
state. Figure~\ref{fig:unemployment_clusters} shows time series for 5
states, along with national unemployment rate and real Gross Domestic
Product (GDP). Typical analyses of raw macroeconomic time series are
built on assumptions about temporal dependence and incorporate
multiple model-based smoothing techniques; see
e.g. \citep{blsmethods}. Cyclical macroeconomic dynamics, such as the
business cycle, are demarcated using additional formal and informal
techniques for assessing agreement across multiple indicators
\citep{nbercycle}. For this analysis, the input data was organized as
a table, where each row $r$ represents a state, each column $c$
represents a month, and each cell $x_{r,c}$ represents the
unemployment rate for state $r$ in month $c$. This representation
removes all temporal cues.

\begin{figure}[t]
\begin{center}
(a) \includegraphics[width=5.7in]{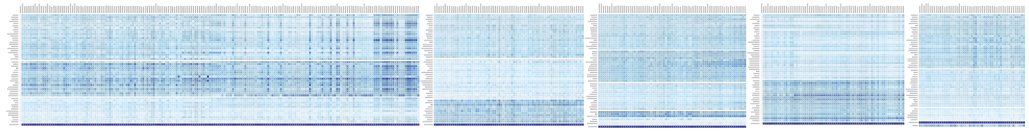} \\
(b) \includegraphics[width=5.7in]{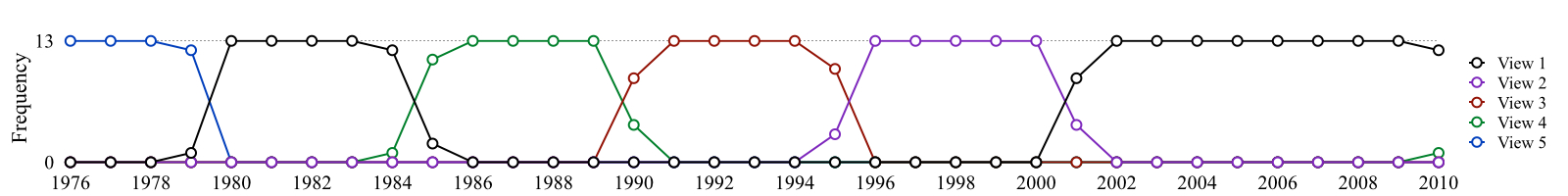} \\
\caption{{\bf The temporal structure in a single posterior sample.}
(a) The complete state-level monthly employment rate dataset, sorted according to one posterior sample. This sample divides the months into 5 time periods, each inducing a different clustering of the states. (b) The frequency of years and quarters for each view shows that each view reflects temporal contiguity: each view either corresponds to a single, temporally contiguous interval or to the union of two such intervals. This temporal structure is not given to CrossCat, but rather inferred from patterns of unemployment across clusters of states.}
\label{fig:unemployment_view}
\end{center}
\end{figure}

CrossCat posits short-range, common-sense temporal structure in
state-level employment rates. The top panel of
Figure~\ref{fig:unemployment_clusters} shows the largest and smallest
month in each of the views in a single posterior sample as vertical
dashes. Figure~\ref{fig:unemployment_view}a shows the frequency of
years in each view; each view contains one or two temporally
contiguous blocks. Figure~\ref{fig:unemployment_view}b shows the raw
unemployment rates sorted according to the cross-categorization from
this sample. Different groups of states are affected by each phase of
the business cycle in different ways, inducing different natural
clusterings of unemployment rates.

\begin{figure}[t]
\begin{center}
\includegraphics[width=5.5in]{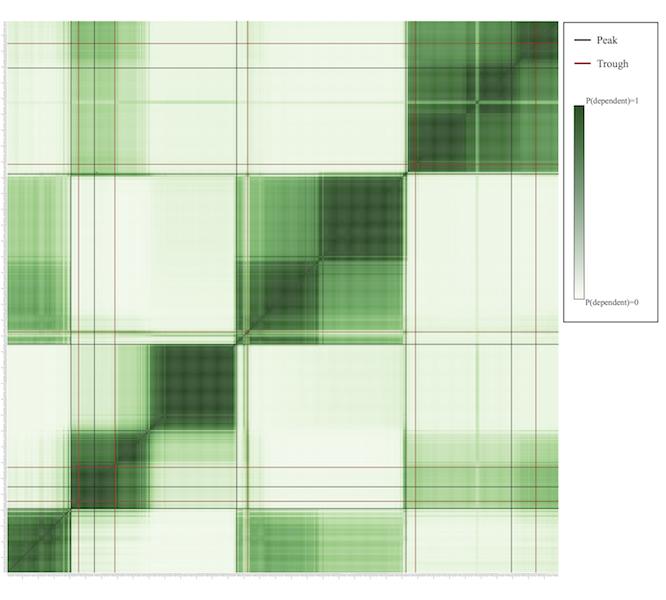}
\caption{
{\bf Dependence probabilities between monthly unemployment rates.} Unemployment rates are sorted by date, beginning with 1976 in the bottom left corner, with business cycle peaks and troughs identified by the NBER in black and red. The beginnings of three major recessions --- in the early 1980s, 1990s, and late 2001 --- are identified by breaks in dependence probability: unemployment rates are dependent with high probability within these periods, and independent of rates from the previous period. See main text for more discussion.}
\label{fig:unemployment_feature-z}
\end{center}
\end{figure}

CrossCat also detects long-range temporal structure that is largely in
agreement with the officially designated phases of the business
cycle. Figure~\ref{fig:unemployment_feature-z} shows the dependence
probabilities for all months, in temporal order, with business cycle
peaks in black and troughs in red. The beginning of the 1980 recession
aligns closely with a sharp drop in dependence probability; this
indicates that during 1980 the states naturally cluster differently
than they do in 1979. Three major US recessions --- 1980, 1990, and
late 2001 --- align with these breakpoints. The beginning of the 2008
recession and the end of the 1980s recession (in 1984) both fall near
sub-block boundaries; these are best seen at high
resolution. Correspondence is not perfect, but this is expected:
CrossCat is not analyzing the same data used to determine business
cycle peaks and troughs, nor is it explicitly assuming any temporal
dynamics. 

Time series analysis techniques commonly assume temporal smoothness
and sometimes also incorporate the possibility of abrupt changes
\citep{ahn1988jump,wang2000bayesian}. CrossCat provides an alternative
approach that makes weaker dynamical assumptions: temporal smoothness
is not assumed at the outset but must be inferred from the data. This
cross-sectional approach to the analysis of natively longitudinal data
may open up new possibilities in econometrics. For example, it could
be fruitful to apply CrossCat to a longitudinal dataset with multiple
macroeconomic variables for each state, or to use CrossCat to combine
temporal information at different timescales.

\FloatBarrier

\section{Discussion}

This paper contains two contributions. First, it describes CrossCat, a
model and inference algorithm that together comprise a domain-general
method for characterizing the full joint distribution of the variables
in a high-dimensional dataset. CrossCat makes it possible to draw a
broad class of Bayesian inferences and to solve prediction problems
without domain-specific modeling. Second, it describes applications to
real-world datasets and analysis problems from multiple
fields. CrossCat finds latent structures that are consistent with
accepted findings as well as common-sense knowledge and that can yield
favorable predictive accuracy compared to generative and
discriminative baselines.

CrossCat is expressive enough to recover several standard statistical
methods by fixing its hyper-parameters or adding other deterministic
constraints:

\begin{enumerate}

\item {\bf Semi-supervised naive Bayes:} $\alpha = 0 \implies \#(unique(\vec{z})) = 1$ and $\alpha_0 = \epsilon$ with $\lambda_{\mathrm{class}} = 0$

  Assume that the dimension in the dataset labeled $class$ contains
  categorical class labels, with a multinomial component model and
  symmetric Dirichlet prior, with concentration parameter
  $\lambda_{\mathrm{class}}$. Because the outer CRP concentration
  parameter is 0, all features as well as the class label will be
  assigned to a single view. Because $\lambda_{\mathrm{class}} = 0$,
  each category in this view will have a component model for the class
  label dimension that constrains the category to only contain data
  items whose class labels all agree:

$$
y_0^i = y_0^j \iff x_{(i,class)} = x_{(j,class)}
$$

This yields a model that is appropriate for semi-supervised
classification, and related to previously proposed techniques based on
EM for mixture models \citep{nigam1999}. Data from each class will be
modeled by a separate set of clusters, with feature hyper-parameters
(e.g. overall scales for continuous values, or levels of noise for
discrete values) shared across classes. Data items whose class labels
are missing will be stochastically assigned to classes based on how
compatible their features are with the features for other data items
in the same class, marginalizing out parameter uncertainty. Forcing
the concentration parameter $\alpha_0$ for the sole inner CRP to have
a sufficiently small value $\epsilon$ ensures that there will only be
a single cluster per class (with arbitrarily high probability
depending on $N$ an d $\epsilon$). These restrictions thus recover a
version of the naive Bayesian classifier (for discrete data) and
linear discriminant analysis (for continuous data), adding
hyper-parameter inference.

\item {\bf Nonparametric mixture modeling:} $\alpha = 0 \implies \#(unique(\vec{z})) = 1$

If the constraints on categorizations are relaxed, but the outer CRP is still constrained to generate a single view, then CrossCat recovers a standard nonparametric Bayesian mixture model. The current formulation of CrossCat additionally enforces independence between features. This assumption is standard for mixtures over high-dimensional discrete data. Mixtures of high-dimensional continuous distributions sometimes support dependence between variables within each component, rather than model all dependence using the latent component assignments. It would be easy and natural to relax CrossCat to support these component models and to revise the calculations of marginal dependence accordingly.

\item {\bf Independent univariate mixtures:} $\alpha = \infty \implies \#(unique(\vec{z})) = D$

The outer CRP can be forced to assign each variable to a separate view by setting its concentration parameter $\alpha$ to $\infty$. With this setting, each customer (variable) will choose a new table (view) with probability 1. In this configuration, CrossCat reduces to a set of independent Dirichlet process mixture models, one per variable. A complex dataset with absolutely no dependencies between variables can induce a CrossCat posterior that concentrates near this subspace.

\item {\bf Clustering with unsupervised feature selection:} $\#(unique(\vec{z})) = 2$, with $\alpha_0 > 0$ but $\alpha_1 = 0$

A standard mixture must model noisy or independent variables using the same cluster assignments as the variables that support the clustering. It can therefore be useful to integrate mixture modeling with feature selection, by permitting inference to select variables that should be modeled independently. The ``irrelevant'' features can be modeled in multiple ways; one natural approach is to use a single parametric model that can independently adjust the entropy of its model for each dimension. CrossCat contains this extension to mixtures as a subspace. 

\end{enumerate}

The empirical results suggest that CrossCat's flexibility in principle
manifests in practice. The experiments show that can effectively
emulate many qualitatively different data generating processes,
including processes with varying degrees of determinism and diverse
dependence structures. However, it will still be important to
quantitatively characterize CrossCat's accuracy as a density
estimator.

Accuracy assessments will be difficult for two main reasons. First, it
is not clear how to define a space of data generators that spans a
sufficiently broad class of applied statistics problems. CrossCat
itself could be used as a starting point, but key statistical
properties such as the marginal and conditional entropies of groups of
variables are only implicitly controllable. Second, it is not clear
how to measure the quality of an emulator for the full joint
distribution. Metrics from collaborative filtering and imputation,
such as the mean squared error on randomly censored cells, do not
account for predictive uncertainty. Also, the accuracy of estimates of
joint distributions and conditional distributions can diverge. Thus
the natural metric choice of KL divergence between the emulated and
true joint distributions may be misleading in applications where
CrossCat is used to respond to a stream of queries of different
structures. Because there are exponentially many possible query
structures, random sampling will most likely be needed. Modeling the
likely query sequences or weighting queries based on their importance
seems ultimately necessary but difficult.

In addition to these questions, CrossCat has several known limitations
that could be addressed by additional research:

\begin{enumerate}

\item {\bf Real-world datasets may contain types and/or shapes of data that CrossCat can only handle by transformation.}

  First, several common data types are poorly modeled by the set of
  component models that CrossCat currently supports. Examples include
  timestamps, geographical locations, currency values, and categorical
  variables drawn from open sets. Additional parametric component
  models --- or nonparametric models, e.g. for open sets of discrete
  values --- could be integrated. 

  Second, it is unclear how to best handle time series data or
  panel/longitudinal settings. In the analysis of state-level monthly
  unemployment data, each state was represented as a row, and each
  time point was a separate and a priori independent column. The
  authors were surprised that CrossCat inferred the temporal structure
  rather than under-fit by ignoring it. In retrospect, this is to be
  expected in circumstances where the temporal signal is sufficiently
  strong, such that it can be recovered by inference over the
  views. However, computational and statistical limitations seem
  likely to lead to under-fitting on panel data with sufficiently many
  time points and variables per time point.

  One pragmatic approach to resolving this issue is to transform panel
  data into cross-sectional data by replacing the time series for each
  member of the population with the parameters of a time-series
  model. Separately fitting the time-series model could be done as a
  pre-processing step, or alternated with CrossCat inference to yield
  a joint Gibbs sampler. Another approach would be to develop a
  sequential extension to CrossCat. For example, the inner Dirichlet
  process mixtures could be replaced with nonparametric state machines
  \citep{beal2001infinite}. Each view would share a common state
  machine, with common states, transition models, and observation
  models. Each group of dependent variables would thus induce a
  division of the data into subpopulations, each with a distinct
  hidden state sequence.

\item {\bf Discriminative learning can be more accurate than CrossCat on standard classification and regression problems.}

  Discriminative techniques can deliver higher predictive accuracy
  than CrossCat when input features are fully observed during both
  training and testing and when there is enough labeled training data.
  One possible remedy is to integrate CrossCat with discriminative
  techniques, e.g. by allowing ``discriminative target'' variables to
  be modeled by generic regressions (e.g. GLMs or Gaussian processes).
  These regressions would be conditioned on the non-discriminative
  variables that would still be modeled by CrossCat.

  An alternative approach is to distinguish prediction targets within
  the CrossCat model. At present, the CrossCat likelihood penalizes
  prediction errors in all features equally. This could be fixed by
  e.g. deterministically constraining the Dirichlet concentration
  hyper-parameters for class-label columns to be equal to 0. This
  forces CrossCat to assign 0 probability density to states that put
  items from different classes into the same category in the view
  containing the class label. These purity constraints can be met by
  using categories that either exactly correspond to the
  finest-grained classes in the dataset or subdivide these
  classes. Conditioned on the hyper-parameters, this modification
  reduces joint density estimation to independent nonparametric
  Bayesian estimation of class-conditional densities
  \citep{aclass2007}.

\item {\bf Natural variations are challenging to test due to the cost
    and difficulty of developing fast implementations.}

  The authors found it surprising that a reliable and scalable
  implementation was possible. Several authors were involved in the
  engineering of multiple high-performance commercial
  implementations. One of these can be applied to multiple real-world,
  million-row datasets with typical runtimes ranging from minutes to
  hours \citep{obermeyer2014scaling}. No fundamental changes to the
  Gibbs sampling algorithm were necessary to make it possible do to
  inference on these scales. Instead, the gains were due to standard
  software performance engineering techniques. Examples include custom
  numerical libraries, careful data structure design, and adopting a
  streaming architecture and compact latent state representation that
  reduce the time spent waiting for memory retrieval. The simplicity
  of the Gibbs sampling algorithm thus turned out to be an asset for
  achieving high performance. Unfortunately, this implementation took
  man-years of software engineering, and is harder to extend and
  modify than slower, research-oriented implementations.

  Probabilistic programming technology could potentially simplify the
  process of prototyping variations on CrossCat and incrementally
  optimizing them.  For example, Venture \citep{mansinghka2014venture}
  can express the CrossCat model and inference algorithm from this
  paper in $\sim$40 lines of probabilistic code. At the time of
  writing, the primary open-source implementation of CrossCat is
  $\sim$4,000 lines of C++. New datatypes, model variations, and
  perhaps even more sophisticated inference strategies could
  potentially be tested this way. However, the performance engineering
  will still be difficult. As an alternative to careful performance
  engineering, the authors experimented with more sophisticated
  algorithms and initializations. Transition operators such as the
  split-merge algorithm from \citep{jainn00splitmerge} and
  initialization schemes based on high-quality approximations to
  Dirichlet process posteriors \citep{liShafto11} did not appear to
  help significantly. These complex approaches are also more difficult
  to debug and optimize than single-site Gibbs. This may be a general
  feature: reductions in the total number of iterations can easily be
  offset by increases in the computation time required for each
  transition.

\end{enumerate}

There is a widespread need for statistical methods that are effective
in high dimensions but do not rely on restrictive or opaque
assumptions \citep{napmassive, wassermanlow}. CrossCat attempts to
address these requirements via a divide-and-conquer strategy. Each
high-dimensional modeling problem is decomposed into multiple
independent subproblems, each of lower dimension. Each of these
subproblems is itself decomposed by splitting the data into discrete
categories that are separately modeled using parametric Bayesian
techniques. The hypothesis space induced by these stochastic
decompositions contains proxies for a broad class of data generators,
including some generators that are simple and others that are
complex. The transparency of simple parametric models is largely
preserved, without sacrificing modeling flexibility. It may be
possible to design other statistical models around this algorithmic
motif.

CrossCat formulates a broad class of supervised, semi-supervised, and
unsupervised learning problems in terms of a single set of models and
a single pair of algorithms for learning and prediction. The set of
models and queries that can be implemented may be large enough to
sustain a dedicated probabilistic programming language. Probabilistic
programs in such a language could contain modeling constraints,
translated into hyper-parameter settings, but leave the remaining
modeling details to be filled in via approximate Bayesian inference.
Data exploration using CrossCat samples can be cumbersome, and would
be simplified by a query language where each query could reference
previous results.

This flexibility comes with costs, especially in applications where
only a single repeated prediction problem is important. In these
cases, it can be more effective to use a statistical procedure that is
optimized for this task, such as a discriminative learning
algorithm. It seems unlikely that even a highly optimized CrossCat
implementation will be able to match the performance of best-in-class
supervised learning algorithms when data is plentiful and all features
are fully observed.

However, just as with software, sophisticated optimizations also come
with costs, and can be premature. For example, some researchers have
suggested that there is an ``illusion of progress'' in classifier
technology \citep{hand2006classifier} in which algorithmic and
statistical improvements documented in classification research papers
frequently do not hold up in practice. Instead, classic methods seem
to give the best and most robust performance. One interpretation is
that this is the result of prematurely optimizing based on particular
notions of expected statistical accuracy. The choice to formalize a
problem as supervised classification may similarly be premature. It is
not uncommon for the desired prediction targets to change after
deployment, or for the typical patterns of missing values to shift. In
both these cases, a collection of CrossCat samples can be used
unmodified, while supervised methods need to be retrained.

It is unclear how far this direct Bayesian approach to data analysis
can be taken, or how broad is the class of data generating processes
that CrossCat can emulate in practice. Some statistical inference
problems may be difficult to pose in terms of approximately Bayesian
reasoning over a space of proxy generators. Under-fitting may be
difficult to avoid, especially for problems with complex couplings
between variables that exceed the statistical capacity of fully
factored models. Despite these challenges, our experiences with
CrossCat have been encouraging. It is fortunate that, paraphrasing
\citet{box1979robustness}, the statistical models that CrossCat
produces can be simplistic yet still flexible and useful. We thus hope
that CrossCat proves to be an effective tool for the analysis of
high-dimensional data. We also hope that the results in this paper
will encourage the design of other fully Bayesian, general-purpose
statistical methods.

\bibliography{bib}
\end{document}